\RequirePackage{letltxmacro}
\documentclass[pdflatex,sn-mathphys]{sn-jnl}
\usepackage{amsmath}
\jyear{2024}%
\usepackage[utf8]{inputenc}
\usepackage[T5]{fontenc}
\usepackage{tabulary}
\usepackage{latexsym}
\usepackage{times}
\usepackage{float}
\usepackage{makecell}
\usepackage{array,booktabs,arydshln,xcolor}
\usepackage{subcaption,booktabs}
\usepackage{microtype}
\usepackage{multirow}
\usepackage{multicol}
\usepackage{threeparttable} 
\usepackage{tablefootnote}
\usepackage{pifont}
\usepackage{caption}
\usepackage{placeins}
\usepackage{array}
\usepackage{textcomp}
\usepackage{lmodern}
\usepackage{graphicx,graphics}

\usepackage{xcolor}
\usepackage{tabularx}
\usepackage{xurl}
\usepackage{amssymb}
\usepackage{pifont}
\usepackage{caption}
\usepackage{algorithm}
\usepackage{algpseudocode}
\usepackage{subcaption}
\usepackage{soul}
\sethlcolor{yellow}

\definecolor{lavender}{RGB}{240,230,250}
\definecolor{lightblue}{RGB}{173, 216, 230}

\definecolor{britishracinggreen}{rgb}{0.0, 0.26, 0.15}
\definecolor{ao(english)}{rgb}{0.0, 0.5, 0.0}
\definecolor{applegreen}{rgb}{0.55, 0.71, 0.0}
\definecolor{lightred}{RGB}{255, 200, 200}

\theoremstyle{thmstyleone}%
\theoremstyle{thmstyletwo}%

\theoremstyle{thmstylethree}%

\begin{document}

\title[ViLegalNLI: Natural Language Inference for
Vietnamese Legal Texts]{ViLegalNLI: Natural Language Inference for Vietnamese Legal Texts}

\author[1,2]{\fnm{Nhung} \sur{Thi-Hong Duong}}\email{22521056@gm.uit.edu.vn}

\author[1,2]{\fnm{Mai} \sur{Ngoc Ho}}\email{22520839@gm.uit.edu.vn}

\author*[1,2]{\fnm{Tin} \sur{Van Huynh}}\email{tinhv@uit.edu.vn}

\author[1,2]{\fnm{Kiet} \sur{Van Nguyen}}\email{kietnv@uit.edu.vn}

\affil[1]{\orgname{University of Information Technology}, \orgaddress{ \city{Ho Chi Minh City}, \country{Vietnam}}}

\affil[2]{\orgname{Vietnam National University}, \orgaddress{ \city{Ho Chi Minh City}, \country{Vietnam}}}


\abstract{In this article, we introduce ViLegalNLI, the first large-scale Vietnamese Natural Language Inference (NLI) dataset specifically constructed for the legal domain. The dataset consists of 42,012 premise–hypothesis pairs derived from official statutory documents and annotated with binary inference labels (Entailment and Non-entailment). It covers multiple legal domains and reflects realistic legal reasoning scenarios characterized by structured logic, conditional clauses, and domain-specific terminology. To construct ViLegalNLI, we propose a semi-automatic data generation framework that integrates large language models for controlled hypothesis generation and systematic quality validation procedures. The framework incorporates artifact mitigation strategies and cross-model validation to improve annotation reliability and ensure legal consistency. The resulting dataset captures diverse reasoning patterns, including paraphrasing, logical implication, and legally invalid inferences, thereby providing a comprehensive benchmark for Vietnamese legal inference tasks. We conduct extensive experiments on the ViLegalNLI using multilingual models, Vietnamese-specific pretrained language models, and instruction-tuned large language models. The results show that few-shot LLM configurations consistently achieve superior performance, while performance is significantly influenced by hypothesis length, lexical overlap, and reasoning complexity. Cross-domain evaluations further reveal the challenges of generalizing legal inference across distinct legal fields. Overall, ViLegalNLI establishes a foundational benchmark for Vietnamese legal NLI and supports future research in legal reasoning, statutory text understanding, and the development of reliable AI systems for legal analysis and decision support. The dataset is publicly available for research purposes. \footnote{Dataset link will be updated upon paper acceptance.}
}


\keywords{Natural Language Inference, Vietnamese Legal Text, Legal Reasoning, Benchmark Dataset, Large Language Models}

\maketitle
\section{Introduction}\label{Introduction}

Natural Language Inference (NLI), also known as Recognizing Textual Entailment (RTE) \cite{MacCartney2009}, is a fundamental task in Natural Language Processing that aims to determine whether a \textit{hypothesis} can be logically inferred from a given \textit{premise} \cite{Dagan2004}. As a widely used benchmark for evaluating natural language understanding, NLI is commonly used to assess whether models can capture semantic meaning and infer logical relationships between sentences.

In the legal domain, NLI is particularly important because legal reasoning often requires determining whether a claim or argument can be derived from statutory provisions or contractual clauses. Legal texts typically exhibit hierarchical structures (e.g., articles, clauses, and points), conditional logic, cross-references, and specialized terminology. Consequently, applying NLI to legal texts requires precise semantic interpretation and structured reasoning that goes beyond simple lexical matching.

Recent advances in NLI have demonstrated its potential for supporting legal language understanding and downstream legal AI applications. In particular, NLI has been applied to tasks such as statutory interpretation, contract analysis, and automated compliance verification, where systems must determine whether a legal statement is supported by relevant legal provisions. Datasets such as ContractNLI \cite{Yuta2021} and LawngNLI  \cite{bruno-roth-2022-lawngnli} have shown that NLI can effectively model reasoning over complex legal texts, enabling systems to identify supporting or contradicting clauses within legal documents.

Despite these advances, research in legal NLI remains largely concentrated on English resources. In Vietnam, the lack of standardized benchmark datasets further limits progress in developing and evaluating legal reasoning models. Although statutory documents are publicly available, they have not been systematically curated and annotated for inference tasks. As a result, the absence of high-quality datasets makes it difficult to rigorously assess the ability of models to perform legal reasoning in Vietnamese.

To address this gap, we introduce \textbf{ViLegalNLI}, the first large-scale Vietnamese dataset for legal natural language inference. The dataset comprises 42,012 premise–hypothesis pairs derived from officially promulgated statutory documents, each annotated with a binary label: \textit{Entailment} or \textit{Non-entailment}. This formulation reflects the practical objective of verifying whether legal claims are supported by statutory provisions.

ViLegalNLI is constructed through a semi-automatic framework that leverages large language models for controlled hypothesis generation and multi-stage validation. The pipeline integrates cross-model label verification and artifact mitigation to improve annotation reliability while maintaining scalability and reducing annotation costs. 

We benchmark multilingual pretrained models, Vietnamese-specific transformer models such as PhoBERT \cite{Nguyen2020PhoBERT}, viBERT \cite{VietviBERT2020}, and CafeBERT \cite{VLue2024}, as well as instruction-tuned large language models including the Gemma and Qwen families. Through comprehensive experiments, we analyze model performance across inference labels, legal domains, sentence lengths, lexical overlap, and cross-domain settings to better understand the challenges of Vietnamese legal reasoning. 

The main contributions of this paper are threefold:
\begin{itemize}
\item \textbf{ViLegalNLI Dataset:} We introduce ViLegalNLI, the first large-scale Vietnamese legal NLI benchmark comprising 42,012 annotated premise–hypothesis pairs derived from official statutory documents.

\item \textbf{Scalable Construction Framework:} We propose a semi-automatic data generation and validation pipeline that integrates LLM-based hypothesis generation with consensus-driven quality control to ensure annotation reliability and scalability.

\item \textbf{Comprehensive Evaluation:} We conduct extensive benchmarking of multilingual, Vietnamese-specific, and instruction-tuned models, along with detailed analyses across inference labels, legal domains, and linguistic factors to assess Vietnamese legal reasoning performance.
\end{itemize}

The remainder of this paper is organized as follows. Section~\ref{Related} reviews related work, and Section~\ref{TaskDefinition} defines the task. Section~\ref{Corpus} describes the dataset and the construction framework, while Section~\ref{QAS} presents the experimental setup. Section~\ref{Ex} reports the results, followed by Section~\ref{discussion}, which discusses the findings. Finally, Section~\ref{sec:conclusion_future_work} concludes the paper and outlines future research directions.

\section{Related Works}\label{Related}

\subsection{Related Datasets}
Benchmark datasets have played a pivotal role in the development of Natural Language Inference (NLI), supporting model training, evaluation, and systematic comparison. Early NLI research predominantly focused on English, leading to the creation of several large-scale datasets that established standard evaluation frameworks for the task. Foundational datasets such as SICK \cite{Marelli2014}, SNLI \cite{Bowman2015}, and MultiNLI \cite{Williams2018} significantly shaped the development of NLI research. Domain-specific datasets, including MedNLI \cite{romanov2018}, SciTail \cite{Khot2018}, and ContractNLI \cite{Yuta2021}, further demonstrated the applicability of NLI to specialized domains such as healthcare, science, and law. Moreover, ANLI \cite{Nie2020}, constructed using an adversarial data collection paradigm, exposed critical weaknesses in contemporary NLI models and underscored the challenges of robust reasoning.

To extend NLI research beyond English and evaluate cross-lingual generalization, numerous multilingual and language-specific datasets have been proposed. Prominent examples include XNLI \cite{Conneau2018}, VLSP2021-NLI \cite{JCSCE}, VietX-NLI \cite{11133930},  OCNLI \cite{Hu2020}, IndoNLI \cite{Mahendra2021}, KorNLI \cite{Ham2020}, and FarsTail \cite{amirkhani2023farstail}, as well as datasets developed for mixed-language settings such as Hinglish \cite{khanuja2020new}. These datasets capture diverse linguistic and cultural characteristics  across regions, thereby supporting the development of multilingual and cross-domain inference models. Table \ref{tab:related_datasets} provides a comparative overview of representative NLI datasets in terms of language, text genre, year, and dataset size.

In parallel with global research trends, several Vietnamese NLI datasets have been released in recent years, including ViNLI \cite{Huynh2022}, ViANLI \cite{Huynh2025}, ViHealthNLI \cite{Nguyen2024ViHealthNLI}, and VnNewsNLI \cite{VnNewsNLI}. These datasets span domains such as general text, news, and healthcare, thereby providing a valuable foundation for training and evaluating NLI models in Vietnamese. However, the legal domain remains largely underexplored. Legal language is characterized by formal style, complex syntactic structures, and strict logical reasoning requirements. To date, no Vietnamese NLI dataset has been constructed directly from official legal documents. Existing legal NLI resources, such as ContractNLI \cite{Yuta2021} and LawngNLI \cite{bruno-roth-2022-lawngnli}, are limited to English and do not adequately reflect the linguistic and structural characteristics of the Vietnamese legal system.

This gap highlights the need for a dedicated Vietnamese legal NLI dataset to advance research in legal language understanding. Such a dataset would also support the development of practical AI systems for legal analysis and decision support.
\begin{table}[]
\centering
\caption{Related Natural Language Inference Datasets.}
\label{tab:related_datasets}
\resizebox{1\linewidth}{!}
{
\begin{tabular}{lcccc}
\hline
\textbf{Dataset}
& \textbf{Language}
& \textbf{Text Genre}
& \textbf{Year}
& \textbf{Quantity} \\
\hline
\multicolumn{5}{c}{\textit{English Datasets}} \\
\hline
SICK \cite{Marelli2014}   & English & Image \& video captions & 2014 & \textasciitilde
10K \\
SNLI \cite{Bowman2015}  & English & Multi-genre & 2018 & \textasciitilde
433K \\
MedNLI \cite{romanov2018}          & English & Medical & 2019 & \textasciitilde
14K \\
SciTail \cite{Khot2018} & English & Education & 2018 & \textasciitilde
27K \\
ANLI \cite{Nie2020} & English & Multi-genre & 2020 & \textasciitilde
169K \\
ContractNLI \cite{Yuta2021} & English & Legal & 2021 & 607 \\
LawngNLI \cite{bruno-roth-2022-lawngnli} & English & Legal & 2022 & 140K \\
\hline
\multicolumn{5}{c}{\textit{Multilingual and Other Language Datasets}} \\
\hline
ArbTEDS \cite{alabbas2013dataset} & Arabic & Newswire & 2013 & 600 \\
OCNLI \cite{Hu2020} & Chinese & Multi-genre & 2020 & \textasciitilde
56K \\
IndoNLI \cite{Mahendra2021} & Indonesian & Multi-genre & 2021 & \textasciitilde
18K \\
KorNLI \cite{Ham2020} & Korean & Multi-genre & 2020 & \textasciitilde
950K \\
FarsTail \cite{amirkhani2023farstail} & Persian & Education & 2021 & \textasciitilde
10K \\
Hinglish \cite{khanuja2020new} & Multi-language & Movie scripts & 2020 & 2.240 \\
XNLI \cite{Conneau2018} & Multi-language & Multi-genre & 2018 & \textasciitilde
129K \\
VLSP2021-NLI \cite{JCSCE} & Multi-language & Newswire & 2022 & \textasciitilde
20K \\
VietX-NLI \cite{11133930} & Multi-language & Newswire & 2025 & \textasciitilde
20K \\
\hline
\multicolumn{5}{c}{\textit{Vietnamese Datasets}}\\
\hline
ViNLI \cite{Huynh2022} & Vietnamese & Newswire & 2022 & \textasciitilde
30K \\
ViANLI \cite{Huynh2025} & Vietnamese & Newswire & 2025 & \textasciitilde
10K \\
ViHealthNLI \cite{Nguyen2024ViHealthNLI} & Vietnamese & Medical & 2024 & \textasciitilde
19K \\
VnNewsNLI \cite{VnNewsNLI} & Vietnamese & Newswire & 2022 & \textasciitilde
32K \\
\hline
\textbf{ViLegalNLI (Ours)} & \textbf{Vietnamse} & \textbf{Legal} & \textbf{2026} & \textbf{\textasciitilde
42K} \\
\hline 
\end{tabular}
}
\end{table}

\subsection{Related Natural Language Inference Model}
The progress of Natural Language Inference (NLI) has been closely driven by the evolution of Transformer-based models. In contrast to traditional approaches based on handcrafted features or rule-based reasoning, modern architectures learn semantic and inferential relations between premise–hypothesis pairs through large-scale pretraining and task-specific fine-tuning. This paradigm enables models to capture complex semantic phenomena, including implicit reasoning, logical contradictions, and contextual dependencies. Existing research on NLI models can be broadly categorized into four directions: multilingual models, enhanced Transformer architectures, Vietnamese monolingual models, and large language models (LLMs).

In the multilingual setting, mBERT \cite{Devlin2019} was among the earliest models to facilitate cross-lingual inference through joint pretraining on over 100 languages. However, its effectiveness is limited for languages with distinct syntactic and lexical properties, such as Vietnamese. XLM-R \cite{Conneau2020} addressed these limitations by leveraging substantially larger multilingual corpora derived from CommonCrawl. This leads to improved cross-lingual generalization and its widespread adoption as a backbone model for multilingual NLI tasks. Building on this foundation, InfoXLM \cite{Chi2021} further improves cross-lingual transfer by integrating masked language modeling with contrastive learning objectives, enabling more effective alignment across languages and demonstrating strong performance in cross-lingual and zero-shot scenarios, particularly for low-resource languages.

In addition to multilingual pretraining, architectural improvements to Transformers have also contributed to NLI performance gains. DeBERTa \cite{He2020DeBERTa}, for instance, introduces a disentangled attention mechanism that separately models content and positional information, enhancing the representation of complex logical relations. 

In addition, Vietnamese monolingual models such as PhoBERT \cite{Nguyen2020PhoBERT}, viBERT \cite{VietviBERT2020}, and domain-adapted variants including CafeBERT \cite{VLue2024} aim to better capture language-specific characteristics. These models often yield improved performance when the pretraining data closely aligns with the target domain.

More recently, large language models (LLMs) have introduced new paradigms for NLI through both fine-tuning and prompting-based inference. Despite their strong general reasoning capabilities, existing evaluations of LLMs predominantly focus on general-domain benchmarks. This raises important questions regarding their effectiveness in low-resource settings and specialized domains such as Vietnamese legal texts, underscoring the need for systematic investigation.

\section{Task Definition}
\label{TaskDefinition}
In the legal domain, Natural Language Inference (NLI) aims to determines the relationship between a legal statement and a statutory provision. The goal is to determine whether the statement is supported by the legal text. This capability is essential for statutory interpretation, legal claim validation, contradiction detection, and automated legal decision support.

In this study, we formulate Vietnamese legal NLI as a binary classification task over sentence pairs. Each pair consists of a \textit{premise} and a \textit{hypothesis}. The model predicts whether the hypothesis can be inferred from the premise within a legal context.

Formally, the task is defined as follows:

\textbf{Input:} A sentence pair $(p, h)$ where:
\begin{itemize}
    \item \textit{Premise} ($p$): A sentence extracted from officially promulgated Vietnamese statutory documents (e.g., articles, clauses, or points currently in effect). The premise serves as the legal basis for inference.
    \item \textit{Hypothesis} ($h$): A constructed statement designed to evaluate the model's reasoning capability. The hypothesis may paraphrase, generalize, specialize, or contradict information stated in the premise.
\end{itemize}

\textbf{Output:} A binary inference label $y \in \{0,1\}$ defined as:
\begin{itemize}
    \item \textit{Entailment} (1): The hypothesis can be logically inferred from the premise or is semantically equivalent in the legal context.
    \item \textit{Non-entailment} (0): The hypothesis cannot be inferred from the premise, including cases of contradiction, semantic inconsistency, or insufficient legal support.
\end{itemize}

\begin{figure}[h]
    \centering
    \includegraphics[width=0.8\linewidth]{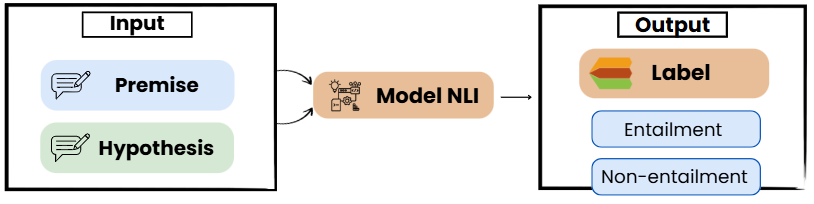}
    \caption{Overview of the Vietnamese legal NLI task. A premise–hypothesis pair is fed into a language model, which predicts a binary inference label (Entailment or Non-entailment).}
    \label{fig:task_overview}
\end{figure}

Figure~\ref{fig:task_overview} illustrates the overall task formulation of the proposed framework, where the premise–hypothesis pair is fed into a language model to predict the corresponding inference label. To further clarify the annotation scheme, Table~\ref{tab:legal_nli_example} presents representative examples from the legal domain.

\begin{table*}[!ht]
\centering
\caption{Illustrative examples of legal NLI}
\resizebox{\textwidth}{!}{
\label{tab:legal_nli_example}
\renewcommand{\arraystretch}{1.1}
\begin{tabular}{p{2cm} p{3.5cm} p{3cm} p{1.5cm}}
\hline
\textbf{Citation} & \textbf{Premise} & \textbf{Hypothesis} & \textbf{Label} \\
\hline

Point a, Clause 2, Article 40, Chapter III, Law No. 59/2024/QH15 (Juvenile Justice)

& Khiển trách có thể được áp dụng đối với người từ đủ 16 tuổi đến dưới 18 tuổi phạm tội nghiêm trọng do vô ý hoặc phạm tội ít nghiêm trọng theo quy định của Bộ luật Hình sự. \textit{(Reprimand shall apply to the following cases: Individuals from 16 to under 18 of age who involuntarily commit very serious crime, or commit a less serious crime as defined by the Criminal Code)}

& Người từ 16 đến dưới 18 tuổi phạm tội nghiêm trọng do vô ý có thể bị áp dụng hình thức khiển trách. \textit{(A person aged from 16 to under 18 who commits a serious crime due to negligence may be subject to a reprimand.)}

& Entailment \\

\hline

Point g, Clause 2, Article 66, Chapter VI, Law No. 54/2024/QH15 (Geology and Minerals)

& Các hành vi bị nghiêm cấm bao gồm: Thông tin không đúng sự thật, kích động, xuyên tạc, phỉ báng về tổ chức và hoạt động công đoàn. \textit{(Mineral production license shall be terminated if: Land or sea waters for mineral production of license holder is revoked in accordance with land laws and other relevant laws where investment project violates land laws or other relevant laws;)}

& Công dân được phép tự do phát biểu và cung cấp thông tin không bị hạn chế về tổ chức công đoàn. \textit{(Citizens are allowed to freely express opinions and provide information without restriction regarding trade union organizations.)}

& Non-entailment \\

\hline
\end{tabular}
}
\end{table*}

\section{Dataset}\label{Corpus}
\subsection{Dataset Creation}
In this section, we describe the dataset construction pipeline shown in Figure~\ref{fig:dataset_pipeline}. 
The process consists of seven steps: 

\begin{figure}[ht]
    \centering
    \includegraphics[width=0.9\linewidth]{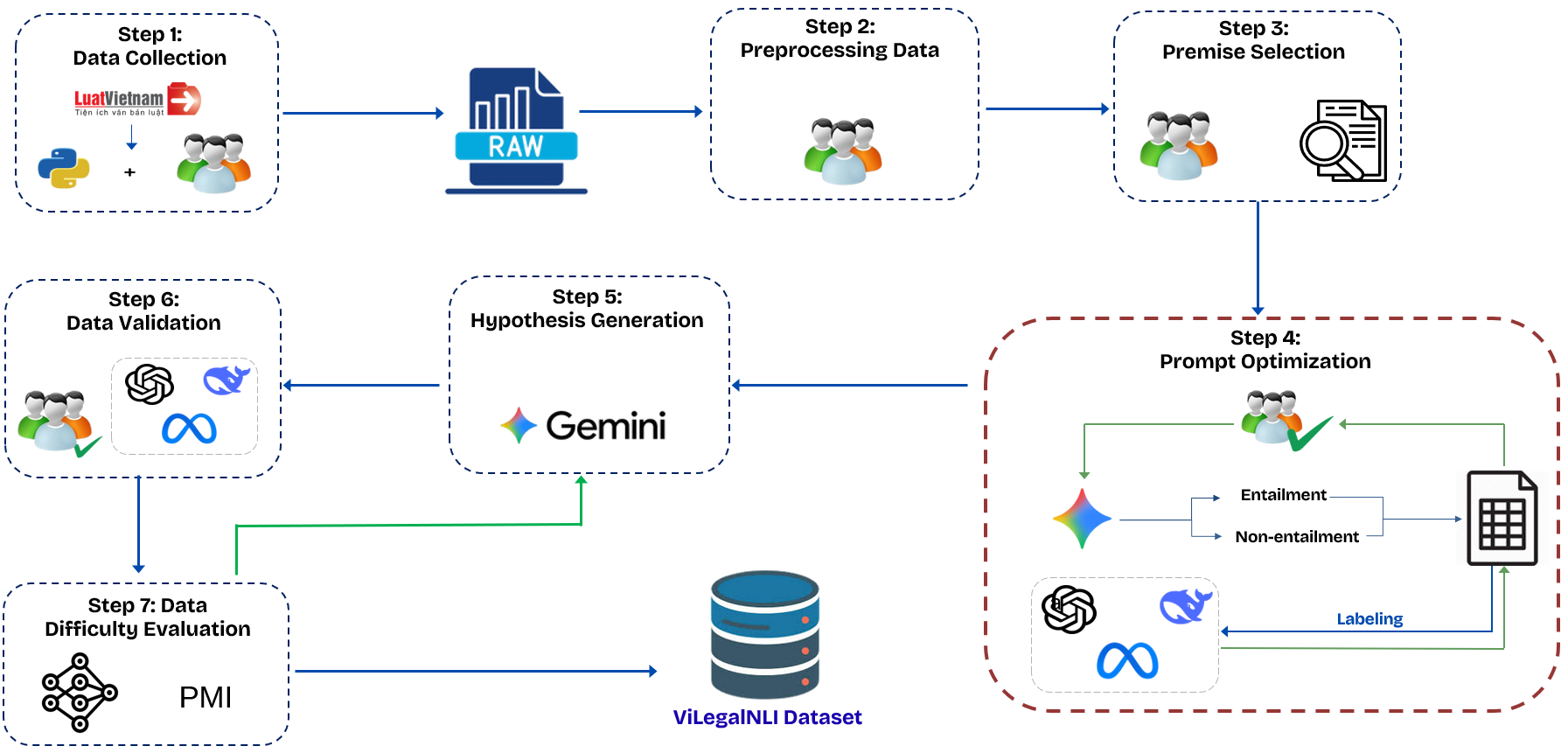}
    \caption{Dataset creation process. This process consists of seven steps: 
(1) Data Collection, 
(2) Data Preprocessing, 
(3) Premise Selection, 
(4) Prompt Optimization and Labeling, 
(5) Hypothesis Generation, 
(6) Data Validation, and 
(7) Data Difficulty Evaluation before forming the final ViLegalNLI dataset.
}
    \label{fig:dataset_pipeline}
\end{figure}

\subsubsection{Data Collection} \label{subsec:data_collection}
To construct a reliable legal corpus for generating premise–hypothesis pairs, we surveyed official legal data sources in Vietnam and selected the Luật Việt Nam portal\footnote{\url{https://luatvietnam.vn/}}
 as the primary data source due to its comprehensive coverage, structured organization, regular updates, and high credibility. These characteristics make it well-suited for legal NLP research.

Data were automatically collected using Selenium in Python to retrieve dynamically generated web content from the portal. A dedicated web crawler was implemented to collect document links and extract full textual content, which was stored in plain-text (.txt) format for subsequent processing. The resulting corpus comprised 514 Vietnamese statutory documents (excluding amendment and supplementary texts) issued up to February 2025.

\subsubsection{Data Preprocessing} \label{subsec:data_preprocessing_cleaning} 

To ensure legal validity and up-to-date relevance, we manually excluded documents that were no longer in effect. The resulting corpus comprised 168 active statutory documents, representing the Vietnamese legal framework as of February 2025.

All documents were subsequently preprocessed to remove noise and standardize formatting. Non-semantic artifacts and decorative separators (e.g., “-----”, “\_\_\_\_\_”, “*****”, “=====”), redundant whitespace, and administrative signatures or confirmation statements were eliminated.

This preprocessing step enhances structural consistency and minimized non-informative artifacts, thereby facilitating more reliable downstream analysis and information extraction.

\subsubsection{Premise Extraction} \label{subsec:premise_extraction}

The premises in ViLegalNLI are automatically extracted from articles, clauses, and sub-clauses of the preprocessed legal documents. As a sentence-level NLI dataset, each premise represents a semantically complete and contextually grounded legal statement derived from statutory provisions.

When clauses constitute the smallest structural units, premises are derived directly from clause-level content. To preserve legal context, article-level information is incorporated when necessary, particularly for provisions describing prohibited acts, sanctions, or semantically interdependent enumerations. For clauses comprising multiple points, premise construction incorporates point-level content. Clause-level context is added when necessary to ensure semantic completeness and contextual coherence. 

\begin{figure}
    \centering
    \includegraphics[width=1\linewidth]{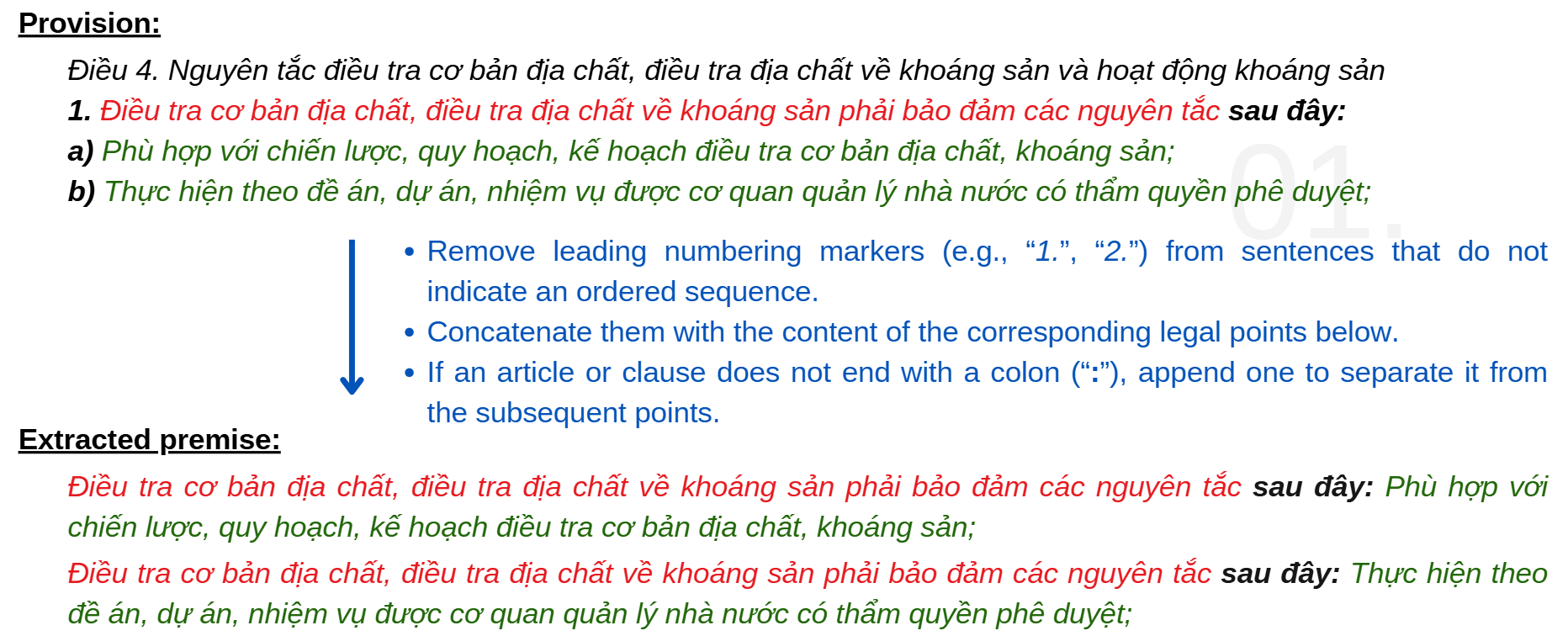}
    \caption{Example of the premise extraction procedure from Vietnamese statutory texts}
    \label{fig:premise_extraction}
\end{figure}

The extraction procedure is implemented automatically using rule-based structural patterns (Table~\ref{tab:rule_based_keyword}). Metadata, including the positions of articles, clauses, and points, as well as document-level attributes (Table \ref{tab:data_attributes}), is simultaneously recorded for traceability. This process yields 20,860 legal premise instances. Figure~\ref{fig:premise_extraction} presents an illustrative example of the extraction workflow.

\begin{table}[ht]
\caption{Rule-based keyword indicators for legal premise extraction}
\resizebox{\textwidth}{!}{
\centering
\label{tab:rule_based_keyword}
\begin{tabular}{p{0.8cm} p{8cm} p{4cm}}
\hline
No. & Keyword Indicators (Vietnamese -- English) & Position within the Provision \\
\hline
1 & ``sau đây'' (\textit{as follows}) / ``sau đây:'' (\textit{as follows:}) & End of the article \\
2 & ``như sau'' (\textit{as follows}) / ``như sau:'' (\textit{as follows:}) & End of the article \\
3 & ``bao gồm'' (\textit{including}) / ``bao gồm:'' (\textit{including:}) & End of the article \\
4 & ``nghiêm cấm'' (\textit{strictly prohibited}), ``cấm'' (\textit{prohibited}), ``vi phạm'' (\textit{violation}), ``bị xử phạt'' (\textit{subject to penalty}), ``không được'' (\textit{not allowed to}) & Within the article \\
5 & ``Thứ tự'' (\textit{order}), ``Trình tự'' (\textit{procedure}) & Within the article \\
\hline
\end{tabular}
}
\end{table}

\begin{table}[ht]
\centering
\caption{Data attributes of the ViLegalNLI dataset}
\label{tab:data_attributes}
\begin{tabular}{clp{7cm}}
\hline
\textbf{No.} & \textbf{Attribute} & \textbf{Description} \\
\hline
1  & Law ID & Official identifier of the legal document \\
2  & Law Name & Title of the legal document \\
3  & Law Date & Date of promulgation \\
4  & Field of Law & Legal sub-domain of the document \\
5  & Chapter & Chapter number \\
6  & Section & Section number \\
7  & Article & Article number corresponding to the premise \\
8  & Clause & Clause number corresponding to the premise \\
9  & Point & Point number corresponding to the premise\\
10 & Premise & Premise text \\
11 & Hypothesis & Hypothesis text \\
12 & Label & Binary inference label indicating the relationship between the premise and hypothesis \\
\hline
\end{tabular}
\end{table}

\subsubsection{Prompt Optimization}
\label{sec:prompt_design}

The ViLegalNLI dataset is constructed using a semi-automatic framework supported by large language models (LLMs). Gemini-2.5 Flash is employed as the primary model for hypothesis generation, while GPT-4o, DeepSeek-R1, and LLaMA-4 Scout are used for cross-model labeling. These models have demonstrated effectiveness in Vietnamese language understanding and are capable of handling complex legal reasoning tasks.

Prompt design is critical for ensuring hypothesis quality and label consistency. 
An initial instruction-based prompt is used to guide Gemini in generating hypotheses under two inference relations: \textit{Entailment} and \textit{Non-entailment}. Independent prompts are designed for the labeling models to maintain evaluation objectivity. Each prompt version was tested on 50 samples per round and iteratively refined based on linguistic clarity and inter-model agreement.

We adopted Fleiss’ Kappa ($\kappa$) to measure inter-model agreement, as it is suitable for multiple annotators. Fleiss’ Kappa evaluates observed agreement ($\bar{P}$) while correcting for chance agreement ($\bar{P}_e$):

\begin{equation}
\kappa = \frac{\bar{P} - \bar{P}_e}{1 - \bar{P}_e}
\end{equation}

Following standard interpretation guidelines, prompts configurations achieving $\kappa \geq 0.85$ were selected for large-scale data generation, corresponding to near-perfect agreement.

\begin{figure}[ht]
    \centering
    \includegraphics[width=0.8\linewidth]{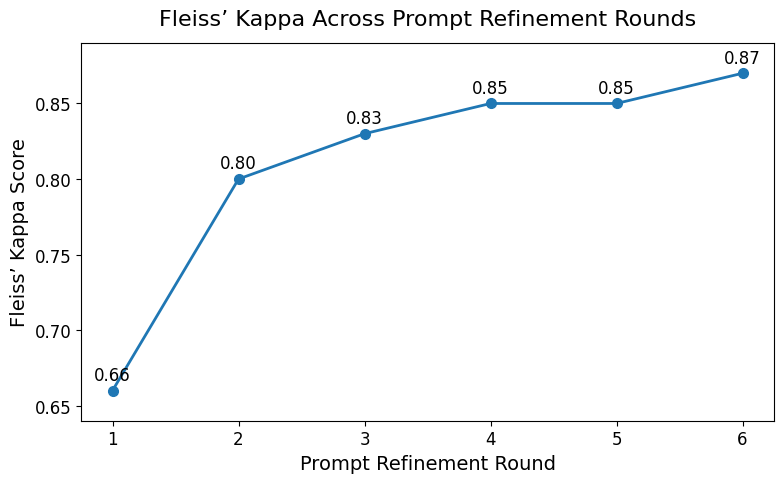}
    \caption{Fleiss’ Kappa across prompt refinement rounds}
    \label{fig:fleiss_kappa}
\end{figure}

Table~\ref{tab:kappa_rounds} summarizes the agreement results across six refinement rounds, while Figure~\ref{fig:fleiss_kappa} presents a visual representation of this progression. The steady increase in Fleiss’ Kappa indicates a positive relationship between prompt refinement and label reliability. The final prompt supports more complex multi-clause reasoning while maintaining high inter-model consistency. This iterative prompt engineering strategy, combining quantitative agreement measurement and qualitative analysis, is effective for constructing a high-quality Vietnamese legal NLI dataset.

\begin{table}[h]
\centering
\caption{Fleiss’ Kappa across prompt refinement rounds}
\label{tab:kappa_rounds}
\renewcommand{\arraystretch}{1.2}
\begin{tabular}{c c l}
\hline
\textbf{Round} & \textbf{$\kappa$} & \textbf{Main Improvement} \\
\hline
1 & 0.67 & Basic prompt, ambiguous hypotheses \\
2 & 0.80 & Added clearer inference constraints \\
3 & 0.83 & Reduced linguistic ambiguity \\
4 & 0.85 & Introduced causal and purposive reasoning \\
5 & 0.85 & Increased diversity of legal scenarios \\
6 & 0.87 & Multi-clause legal reasoning \\
\hline
\end{tabular}
\end{table}

\subsubsection{Hypothesis Generation} \label{subsec:hypothesis _generation}
The hypotheses in ViLegalNLI are automatically generated using the Gemini-2.5 Flash model, guided by a predefined set of generation rules (Table~\ref{tab:generation_rules}) and a finalized prompt derived through iterative refinement (Section \ref{sec:prompt_design}).

The generation framework is specifically designed to reflect the linguistic structure and normative characteristics of Vietnamese legal texts. It is designed to ensure that the generated hypotheses accurately represent standard inference labels (\textit{Entailment} and \textit{Non-entailment}) while preserving the logical consistency and formal structure of statutory provisions.

The prompting strategy follows a structured reasoning paradigm to encourage step-by-step inferential consistency, thereby enhancing logical coherence and label reliability.

\begin{table}[ht]
\centering
\caption{Generation rules for entailment and non-entailment}
\label{tab:generation_rules}
\renewcommand{\arraystretch}{1.1}
\begin{tabular}{
c >{\raggedright\arraybackslash}p{4.7cm} |
c >{\raggedright\arraybackslash}p{4.7cm}
}
\hline
\multicolumn{2}{c|}{\textbf{Entailment}} & 
\multicolumn{2}{c}{\textbf{Non-entailment}} \\
\hline
\textbf{ID} & \textbf{Description} & 
\textbf{ID} & \textbf{Description} \\
\hline
1 & Active–passive transformation with semantic preservation. &
1 & Structural transformation introducing contradiction. \\

2 & Replace entities with synonymous legal references. &
2 & Alter entities, time, or actions. \\

3 & Nominal–clausal reformulation. &
3 & Create semantic inconsistency. \\

4 & Conditional reformulation preserving logic. &
4 & Add contradictory conditions. \\

5 & Equivalent numerical modification. &
5 & Modify numerical values. \\

6 & Sentence complexity expansion without meaning change. &
6 & Generate invalid reasoning statements. \\

7 & Logical consequence inference. &
7 & Introduce unsupported assumptions. \\

8 & General-to-specific rule application. &
8 & Misapply general legal rules. \\

9 & Link to related clauses without altering meaning. &
9 & Link to unrelated clauses. \\

& &
10 & Create independent, non-inferable statements. \\

\hline
\end{tabular}
\end{table}

\subsubsection{Data Validation} \label{subsec:data_evaluation}
After large-scale generation, ViLegalNLI undergoes a dedicated evaluation and label validation phase (Step 6 in Figure ~\ref{fig:dataset_pipeline}) to reduce single-model bias and ensure label reliability.

We adopt a multi-layer evaluation framework that combines automatic cross-labeling with manual verification. Three independent models—GPT-4o, DeepSeek-R1, and LLaMA-4 Scout—re-annotated the entire dataset without access to the original labels generated by Gemini-2.5 Flash, ensuring objective assessment.

A sample is retained if at least two of the three models agree with the original label. Samples with no agreement are discarded, whereas those with only one agreeing model are subjected to manual review involving detailed semantic and legal analysis.

\begin{table}[h]
\centering
\caption{Inter-model agreement statistics on the generated dataset}
\label{tab:data_agreement}
\renewcommand{\arraystretch}{1.2}
\begin{tabular}{c c c c}
\hline
\textbf{Label} & \textbf{3 Agreements} & \textbf{$\geq$2 Agreements} & \textbf{1 Disagreement} \\
\hline
0 (Non-entailment) & 93.28\% & 98.81\% & 1.19\% \\
1 (Entailment) & 62.44\% & 84.75\% & 15.25\% \\
\hline
Total & 79.34\% & 92.45\% & 7.55\% \\
\hline
\end{tabular}
\end{table}

As shown in Table~\ref{tab:data_agreement}, 92.45\% of the data achieved agreement from at least two models, and 79.34\% reached full three-model agreement. Most disagreements can be attributed to labeling variation rather than true inferential inconsistency, with genuinely problematic cases accounting for approximately 1.2\% of the dataset.

Overall, the consensus-based filtering strategy helps ensure high label quality while maintaining sufficient dataset scale for legal NLI research.

\subsubsection{Data Difficulty Evaluation} \label{subsec:data_difficulty_evaluation}

Beyond label validation, Step 7 (Section \ref{subsec:data_difficulty_evaluation}) focus on controlling dataset difficulty, enhancing diversity, and mitigating annotation artifacts—cases where models rely on superficial lexical cues rather than genuine semantic reasoning.

To analyze potential lexical artifacts in the dataset, we employ CafeBERT \cite{VLue2024} as a diagnostic model, given its effectiveness in capturing linguistic patterns in Vietnamese text. An NLI model trained on ViLegalNLI achieves nearly 90\% accuracy (Table~\ref{tab:pmi}). However, competitive performance in a hypothesis-only setting (i.e., without access to the premise) suggests the presence of label-correlated lexical patterns, indicating that models may partially rely on surface-level signals rather than deep semantic reasoning.

\begin{table}[htbp]
\caption{Performance of the NLI model on the dataset before artifact mitigation}
\centering
\label{tab:pmi}
\begin{tabular}{lcccc}
\toprule
Model & \multicolumn{2}{c}{Dev Set} & \multicolumn{2}{c}{Test Set} \\
\cmidrule(lr){2-3} \cmidrule(lr){4-5}
 & Accuracy & F1-score & Accuracy & F1-score \\
\midrule
CafeBERT & 89.91\% & 89.78\% & 88.63\% & 88.50\% \\
\bottomrule
\end{tabular}
\end{table}

To further detect annotation artifacts, we compute Pointwise Mutual Information (PMI) between hypothesis tokens and inference labels.

\begin{equation}
PMI(w, y) = \log \frac{P(w, y)}{P(w)P(y)}
\end{equation}

where $P(w, y)$ denotes the joint probability of token $w$ occurring in hypotheses with label $y$, while $P(w)$ and $P(y)$ represent the marginal probabilities of the token and label, respectively. A higher PMI value indicates a stronger association between a token and a specific label, thereby indicating the presence of potential lexical artifacts.

Tokens with high PMI scores are identified as label-indicative artifacts. Instances containing such tokens are revised through controlled paraphrasing, in which the generative model rewrites the hypothesis while preserving its semantic meaning and removing artifact-related cues. Representative artifact tokens and their corresponding synonym substitutions are presented in Table~\ref{tab:trigger_words}.

\begin{table}[h]
\centering
\caption{Trigger expressions and their semantic variations}
\label{tab:trigger_words}
\renewcommand{\arraystretch}{1.2}

\begin{tabularx}{\linewidth}{c p{2.2cm} c X}
\hline
\textbf{No.} & \textbf{Trigger} & \textbf{Relation} & \textbf{Representative Variants} \\
\hline
1 & ``Dù'' (\textit{although}) & Non-entailment & Mặc dù (\textit{although}); Tuy (\textit{though}); Dẫu (\textit{even though}); Mặc cho (\textit{despite}); Tuy rằng (\textit{although}); etc. \\
2 & ``Khi'' (\textit{when}) & Entailment & Một khi (\textit{once}); Ngay khi (\textit{as soon as}); Trong khi (\textit{while}); Giả sử (\textit{if / suppose}); etc. \\
3 & ``Căn cứ'' (\textit{based on}) & Entailment & Dựa theo (\textit{according to}); Dựa vào (\textit{based on}); Dựa trên cơ sở (\textit{on the basis of}); etc. \\
4 & ``không cần'' (\textit{not required}) & Non-entailment & Chẳng cần (\textit{no need}); Không yêu cầu (\textit{not required}); Không nhất thiết (\textit{not necessarily required}); etc. \\
\hline
\end{tabularx}

\end{table}

\subsubsection{Dataset Splitting} \label{subsec:data_splitting}

The ViLegalNLI dataset is divided into training, validation, and test sets with an 8:1:1 ratio. This split is designed to ensure sufficient training data while maintaining balanced label distributions across Entailment and Non-entailment classes.

To prevent potential data leakage, all hypotheses derived from the same premise are assigned to the same subset, so that no premise appears in multiple splits. This design reduces memorization effects and enables fair evaluation of genuine reasoning capability.

Additionally, label and legal sub-domain distributions are controlled to mitigate potential bias toward specific inference types or legal fields. This structured splitting strategy facilitates reliable model training, evaluation, and benchmarking for Vietnamese legal NLI.

\subsection{Dataset Analysis}

\subsubsection{Dataset Overview and Distribution}

\paragraph{Overview.}
ViLegalNLI is a Vietnamese Natural Language Inference (NLI) dataset developed for the legal domain. It comprises 42,012 premise–hypothesis pairs derived from 168 statutory documents currently in force, covering 27 legal sub-domains. To the best of our knowledge, this is the first Vietnamese dataset specifically designed for legal NLI. It serves as both a benchmark for evaluating Vietnamese NLI models and a testbed for assessing legal reasoning capabilities.

\paragraph{Data Structure.}
Each instance is represented as a structured record that includes legal metadata (e.g., document ID, promulgation date, legal domain, hierarchical positions such as chapter, section, article, clause, and point), along with the textual premise, hypothesis, and inference label (Table~\ref{tab:data_attributes}). 
This structured representation facilitates sub-domain analysis and legal traceability.

\paragraph{Sub-domain Distribution.}
ViLegalNLI covers 27 legal domains (Table~\ref{tab:legal_domains}), indicating broad coverage of the Vietnamese legal system. However, the distribution is not uniform. Sub-domains such as \textit{Administrative Organization} and \textit{Natural Resources–Environment} account for larger proportions of the dataset, while \textit{Intellectual Property and Legal Services} contain fewer instances. This imbalance reflects differences in regulatory density and legislative volume across legal fields. 

\begin{table}[ht]
\centering
\caption{Legal domains covered in ViLegalNLI}
\label{tab:legal_domains}
\begin{tabular}{cl}
\hline
\textbf{No.} & \textbf{Legal Domain} \\
\hline
1  & Administrative Organization \\
2  & State Finance \\
3  & Culture – Society \\
4  & Natural Resources – Environment \\
5  & Real Estate \\
6  & Construction – Urban Planning \\
7  & Commercial Law \\
8  & Sports – Health \\
9  & Education \\
10 & Tax – Fees \\
11 & Transportation \\
12 & Labor – Wages \\
13 & Information Technology \\
14 & Investment \\
15 & Enterprise \\
16 & Import – Export \\
17 & Civil Rights \\
18 & Banking – Monetary \\
19 & Legal Services \\
20 & Insurance \\
21 & Procedural Law \\
22 & Administrative Violations \\
23 & Accounting – Auditing \\
24 & Criminal Liability \\
25 & Intellectual Property \\
26 & Securities \\
27 & Other Domains \\
\hline
\end{tabular}
\end{table}

To ensure fair evaluation, domain proportions are preserved across data splits to mitigate domain-specific bias (Table~\ref{tab:dataset_statistics}).

\begin{table*}[t]
\centering
\small
\caption{Overall statistics of ViLegalNLI across legal domains and data splits}
\label{tab:dataset_statistics}
\begin{tabular}{lcccc}
\hline
\textbf{Legal Sub-domain} & \textbf{Train} & \textbf{Dev} & \textbf{Test} & \textbf{Total} \\
\hline
Administrative Organization & 5119 & 980 & 943 & 7042 \\
State Finance & 1800 & 262 & 232 & 2294 \\
Culture – Society & 3785 & 548 & 491 & 4374 \\
Natural Resources – Environment & 5343 & 350 & 289 & 4182 \\
Real Estate & 2261 & 132 & 111 & 2504 \\
Construction – Urban Planning & 1292 & 158 & 135 & 1585 \\
Commercial Law & 2427 & 349 & 320 & 3096 \\
Sports – Health & 1310 & 129 & 112 & 1151 \\
Education & 756 & 132 & 132 & 1020 \\
Tax – Fees & 476 & 108 & 137 & 721 \\
Transportation & 1326 & 91 & 80 & 1497 \\
Labor – Wages & 1380 & 156 & 137 & 1673 \\
Information Technology & 1046 & 169 & 154 & 1369 \\
Investment & 930 & 102 & 76 & 1008 \\
Enterprise & 2493 & 169 & 135 & 2797 \\
Import – Export & 468 & 78 & 73 & 619 \\
Civil Rights & 1760 & 316 & 285 & 2361 \\
Banking – Monetary & 876 & 104 & 79 & 1059 \\
Legal Services & 278 & 52 & 40 & 370 \\
Insurance & 938 & 104 & 89 & 1131 \\
Procedural Law & 2204 & 299 & 264 & 2767 \\
Administrative Violations & 296 & 26 & 24 & 364 \\
Accounting – Auditing & 574 & 78 & 68 & 720 \\
Criminal Liability & 700 & 164 & 167 & 1031 \\
Intellectual Property & 156 & 26 & 25 & 207 \\
Securities & 420 & 26 & 24 & 472 \\
Other Domains & 1448 & 206 & 189 & 1672 \\
\hline
\textbf{Total} & \textbf{34,121} & \textbf{4,160} & \textbf{3,731} & \textbf{42,012} \\
\hline
\textbf{Avg Premise Length} &43.05 & 43.52 & 42.83 & 43.08 \\
\textbf{Avg Hypothesis Length} & 43.05 & 43.52 & 42.83 & 43.08 \\
\hline
\end{tabular}
\end{table*}

\paragraph{Sentence Characteristics.}
The average length of premises is 43.08 tokens, while hypotheses average 43.74 tokens, indicating relatively long and structurally complex legal sentences. 
Table ~\ref{tab:dataset_statistics} presents the overall label distribution.

\subsubsection{Textual Characteristics and Lexical Overlap}

\begin{figure}[ht]
    \centering
    \begin{subfigure}{0.48\linewidth}
        \centering
        \includegraphics[width=\linewidth]{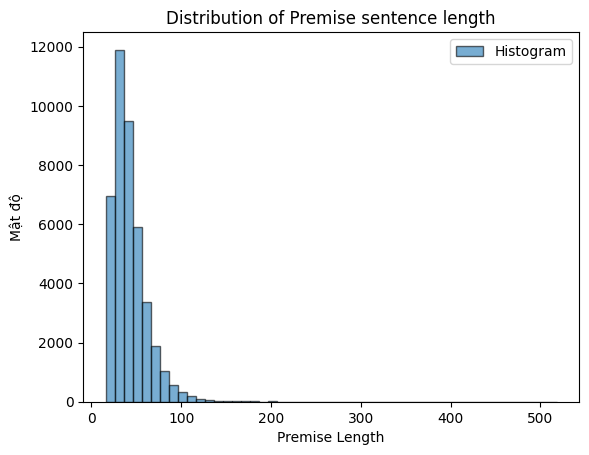}
        \caption{Premise}
        \label{fig:length_pre_distribution}
    \end{subfigure}
    \hfill
    \begin{subfigure}{0.48\linewidth}
        \centering
        \includegraphics[width=\linewidth]{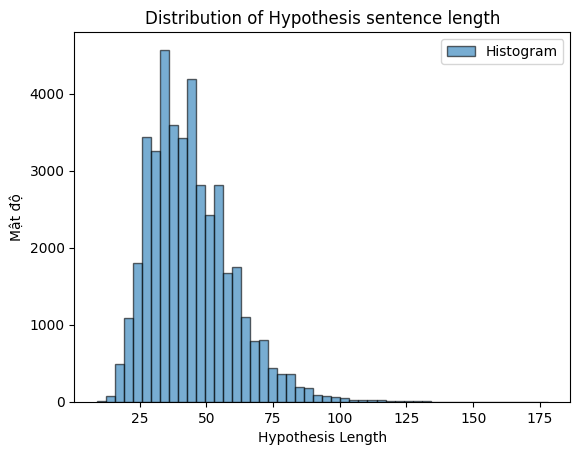}
        \caption{Hypothesis}
        \label{fig:length_hypo_distribution}
    \end{subfigure}
    \caption{Length distribution of premises and hypotheses}
    \label{fig:length_distribution}
\end{figure}

Figure~\ref{fig:length_distribution} shows the length distributions of premises and hypotheses. Premises exhibit a right-skewed distribution, with most sentences ranging between 20–80 tokens and a long tail exceeding 200 tokens. In contrast, hypotheses are more concentrated within 25–60 tokens. This difference suggests that hypotheses are typically concise inferential statements derived from longer premises, reflecting deeper reasoning rather than direct textual paraphrasing.

Lexical overlap is measured using Jaccard similarity, Longest Common Subsequence (LCS), and New Word Rate (Table~\ref{tab:lexical_pos_stats}). Although entailment pairs exhibit slightly higher lexical overlap, both labels maintain substantial new word rates, indicating that hypotheses are not simple surface-level copies of premises.

\begin{table}[ht]
\centering
\caption{Lexical overlap and POS distributions between premise and hypothesis.}
\small
\setlength{\tabcolsep}{2.pt} 
\begin{tabular}{lccc cccccc}
\toprule
\textbf{Label} & \textbf{Jaccard} & \textbf{LCS} & \textbf{New word} & \multicolumn{6}{c}{\textbf{Part-Of-Speech (\%)}} \\
\cmidrule(lr){5-10}
 & (\%) & (\%) & \textbf{rate} (\%) & \textbf{Noun} & \textbf{Verb} & \textbf{Adj} & \textbf{Pron} & \textbf{Adv} & \textbf{Other} \\
\midrule
Entailment      & 28.71 & 41.34 & 60.00 & 32.37 & 26.32 & 3.03 & 2.04 & 4.12 & 0.04 \\
Non-entailment  & 18.43 & 34.29 & 75.39 & 28.85 & 25.47 & 4.15 & 2.40 & 5.87 & 0.09 \\
\bottomrule
\end{tabular}
\label{tab:lexical_pos_stats}
\end{table}

\subsubsection{Linguistic Feature Analysis}

Part-of-speech (POS) distribution of hypotheses is presented in Table~\ref{tab:lexical_pos_stats}. Nouns and verbs dominate both labels, reflecting the entity- and action-centered nature of legal reasoning. A slightly higher proportion of noun in entailment cases suggests stronger alignment with legally defined entities and concepts.

\subsubsection{Generation Rule Distribution}

\begin{figure}[ht]
    \centering
    \includegraphics[width=0.8\linewidth]{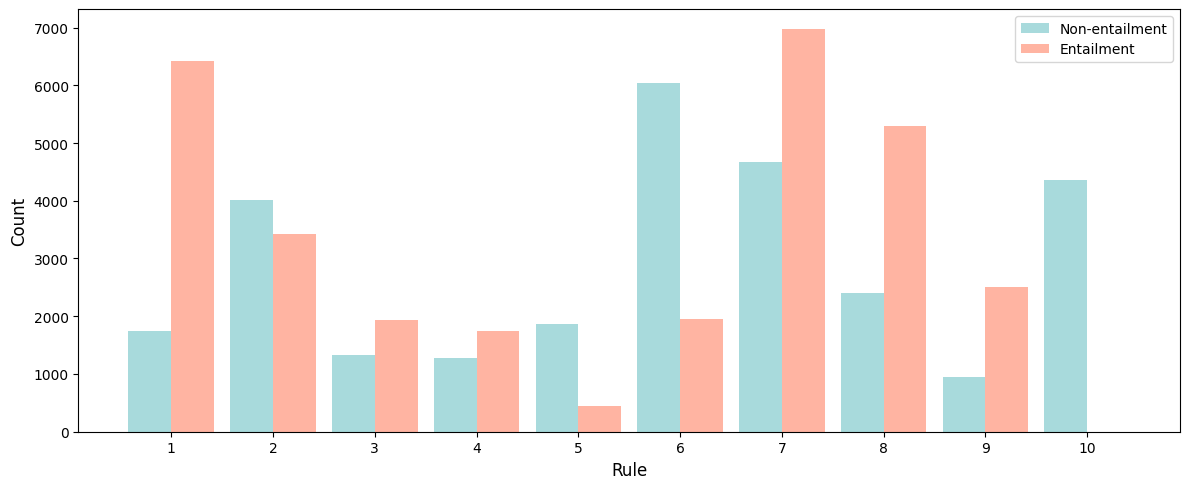}
    \caption{Distribution across generation rules}
    \label{fig:rule_distribution}
\end{figure}

Figure~\ref{fig:rule_distribution} presents the distribution of premise–hypothesis pairs across ten generation rules.

Certain rules (e.g., Rule 1 and Rule 7) tend to yield more entailment pairs, as they involve paraphrasing and valid generalization strategies. In contrast, other rules (e.g., Rule 6 and Rule 10) tend to yield more non-entailment instances through negation or condition reversal. No rule is exclusively associated with a single label, which help reduce rule-based annotation artifacts.

\subsubsection{Artifact Analysis and Dataset Difficulty}

To detect annotation artifacts, we compute Pointwise Mutual Information (PMI) between hypothesis tokens and labels (Table~\ref{tab:pmi_tokens}).

\begin{table}[ht]
\centering
\caption{Top label-discriminative tokens based on PMI}
\label{tab:pmi_tokens}
\begin{tabular}{lccc}
\hline
\textbf{Token} & \textbf{Label} & \textbf{PMI} & \textbf{Frequency (\%)} \\
\hline
không cần (\textit{not required}) & Non-entailment & 0.95 & 13.02 \\
Dù (\textit{although}) & Non-entailment & 0.97 & 8.15 \\
bất kể (\textit{regardless of}) & Non-entailment & 0.89 & 4.26 \\
chỉ cần (\textit{only need}) & Non-entailment & 0.91 & 3.70 \\
tự do (\textit{freely}) & Non-entailment & 0.89 & 3.48 \\
\hline
Khi (\textit{when}) & Entailment & 0.99 & 5.52 \\
Theo quy định (\textit{according to regulations}) & Entailment & 0.98 & 5.44 \\
đồng thời (\textit{simultaneously}) & Entailment & 0.89 & 5.27 \\
Những (\textit{those}) & Entailment & 0.85 & 3.91 \\
thiết lập (\textit{establish / set up}) & Entailment & 0.80 & 2.94 \\
\hline
\end{tabular}
\end{table}

PMI results reveal strong label-specific lexical signals, particularly for Non-entailment. To address this issue, artifact tokens are paraphrased under controlled rewriting constraints, ensuring that superficial lexical cues are minimized.

\begin{figure}[ht]
    \centering
    \includegraphics[width=0.6\linewidth]{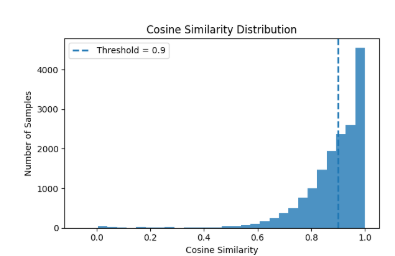}
    \caption{Cosine similarity distribution before and after artifact mitigation}
    \label{fig:cosine_distribution}
\end{figure}

Semantic preservation is verified using cosine similarity between original and revised hypotheses. The average cosine similarity reaches 0.883, with 55.5\% of instances exceeding 0.9. Lower similarity values typically correspond to controlled logical transformations (e.g., negation or scope modification) rather than semantic distortion. Figure~\ref{fig:cosine_distribution} illustrates the cosine similarity distribution.

Overall, these analyses indicate that ViLegalNLI achieves balanced label distribution, broad domain coverage, reduced lexical artifacts, and substantial semantic complexity, making it a challenging and reliable benchmark for legal NLI.

\section{Methodology}\label{QAS}

To evaluate the effectiveness of language inference on the dataset, this study conducts experiments with multiple groups of models, including multilingual pretrained language models, Vietnamese monolingual models, and large language models. The selection of diverse architectures and modeling paradigms is intended to enable a comprehensive comparison of different representation learning and reasoning strategies.
\begin{itemize}

    \item \textbf{Multilingual NLI models} are employed to assess the ability to transfer knowledge from high-resource languages to Vietnamese. Representative models in this group include mBERT \cite{Devlin2019}, XLM-R (Base, Large), and InfoXLM (Base, Large). These encoder-based models are pretrained on large-scale multilingual corpora, allowing them to learn shared semantic representations across languages. This group serves as a baseline for evaluating the effectiveness of language-specific pretraining for Vietnamese.

    \item \textbf{Vietnamese monolingual NLI models} are designed and pretrained specifically for Vietnamese, with the goal of better capturing language-specific characteristics such as compound words, syntactic structures, and semantic expressions. The models used in this group include PhoBERT (Base, Large) \cite{Nguyen2020PhoBERT}, viBERT \cite{VietviBERT2020}, and CafeBERT \cite{VLue2024}. By focusing exclusively on Vietnamese data during pretraining, these models are expected to achieve stronger performance than multilingual models on Vietnamese NLI tasks.

    \item \textbf{Improved Transformer-based models} are included alongside traditional BERT-style architectures. DeBERTa V3 (Base, Large), which represents an enhanced Transformer architecture through its disentangled attention mechanism, is used to examine the impact of architectural improvements on semantic inference performance in Vietnamese.

    \item \textbf{Large Language Models (LLMs)} are incorporated to investigate the inference capabilities of modern decoder-only models in the context of Vietnamese NLI. Two main approaches are considered: fine-tuning a compact LLM (Gemma-2) and applying prompting strategies (zero-shot and few-shot) to more recent models such as Gemma-3 and Qwen2.5. This group enables a direct comparison between inference driven by pretrained knowledge alone and performance gains achieved through task-specific fine-tuning.

\end{itemize}

\section{Experiments and Results}\label{Ex}

\subsection{Experimental Settings}
In this study, we fine-tune the NLI models presented in Section ~\ref{QAS} on the proposed dataset. Evaluation and checkpointing are performed at each epoch, and the best-performing model is selected based on development set performance and subsequently evaluated on the test set.

All models are optimized using a the Adam optimizer with an initial learning rate of 1e-5. The batch size is set to 16 for training and 32 for evaluation, while gradient accumulation with a factor of 2 is applied to stabilize parameter updates. Each model is trained for 5 epochs, with a weight decay of 0.01 to mitigate overfitting.

To improve computational efficiency and reduce memory consumption, mixed-precision training (FP16) is employed. Metrics are logged every 50 steps to monitor training dynamics and convergence behavior. This experimental configuration is kept consistent across all models to ensure fairness and comparability.

\subsection{Evaluation Metrics}

To evaluate model performance on the ViLegalNLI dataset, we adopt two widely used classification metrics: \textit{Accuracy} and macro-averaged \textit{F1-score}.

\paragraph{Accuracy.}
Accuracy measures the proportion of correctly predicted instances over the total number of samples:
\begin{equation}
\text{Accuracy} = \frac{TP + TN}{TP + TN + FP + FN}
\end{equation}
where $TP$, $TN$, $FP$, and $FN$ denote true positives, true negatives, false positives, and false negatives, respectively. While Accuracy provides an overall assessment of model correctness, it may be insufficient in the presence of class imbalance or prediction bias.

\paragraph{F1-score.}
To address this limitation, we additionally report the macro-averaged F1-score (macro-F1), which balances Precision and Recall:
\begin{equation}
\text{F1} = \frac{2 \cdot \text{Precision} \cdot \text{Recall}}{\text{Precision} + \text{Recall}
}
\end{equation}

Macro-F1 is computed by averaging F1-scores across the Entailment and Non-entailment classes, assigning equal weight to each label. This metric provides a more robust evaluation under label imbalance and better reflects the model’s ability to capture both positive and negative inference relations.

Overall, Accuracy and macro-F1 jointly provide a comprehensive evaluation, capturing both overall correctness and balanced reasoning performance in Vietnamese legal NLI.

\subsection{Experimental Results}
Table~\ref{tab:nli_results} reports the performance of a range of models, including multilingual, Vietnamese-specific, and large language models on the Dev and Test sets of ViLegalNLI, evaluated using Accuracy and F1-score. 

In general, large language models in the few-shot setting achieve the strongest performance, with Qwen2.5 obtaining the highest Test accuracy (90.72\%) and F1-score (90.64\%). Among pretrained encoder models, InfoXLM (large) and CafeBERT show strong and stable results, highlighting the importance of domain adaptation and multilingual pretraining. 

In contrast, zero-shot LLMs and smaller monolingual models exhibit noticeably lower performance. This suggests that legal inference remains challenging without task-specific adaptation or in-context guidance. Fine-tuned Gemma-2 improves over zero-shot settings but does not surpass few-shot prompting, indicating that prompt-based adaptation can be highly effective for this task.

\begin{table}[ht]
\centering
\caption{Performance evaluation results of NLI models on the Dev and Test sets.}
\label{tab:nli_results}
\resizebox{1\linewidth}{!}{
\begin{tabular}{lcccc}
\hline
\multirow{2}{*}{\textbf{Model}} 
& \multicolumn{2}{c}{\textbf{Dev Set}} 
& \multicolumn{2}{c}{\textbf{Test Set}} \\
\cline{2-5}
& \textbf{Accuracy (\%)} & \textbf{F1-score (\%)} 
& \textbf{Accuracy (\%)} & \textbf{F1-score (\%)} \\
\hline
\multicolumn{5}{c}{\textit{Multilingual NLI models}} \\
\hline
XLM-R (base)   & 84.34 & 84.18 & 83.65 & 83.43 \\
XLM-R (large)  & 86.98 & 86.81 & 86.37 & 86.19 \\
mBERT          & 80.41 & 79.83 & 79.36 & 78.75 \\
InfoXLM (base) & 83.96 & 83.65 & 83.25 & 82.85 \\
InfoXLM (large) 
& \textbf{89.24} & \textbf{89.15} & \textbf{87.98} & \textbf{87.85} \\
\hline
\multicolumn{5}{c}{\textit{Vietnamese monolingual NLI models}} \\
\hline
PhoBERT (base)  & 85.14 & 84.84 & 84.46 & 84.13 \\
PhoBERT (large) & 86.13 & 85.97 & 84.98 & 84.78 \\
viBERT          & 75.00 & 74.24 & 74.53 & 73.54 \\
CafeBERT        
& \textbf{87.56} & \textbf{87.44} & \textbf{87.49} & \textbf{87.36} \\
\hline
\multicolumn{5}{c}{\textit{Improved Transfomer-based Models}}\\
\hline
DeBERTa V3 (base) & 76.15 & 75.56 & 76.01 & 75.52 \\
\textbf{DeBERTa V3 (large)} & \textbf{86.16} & \textbf{85.98} & \textbf{85.35} & \textbf{85.18} \\
\hline
\multicolumn{5}{c}{\textit{Large Language Models (Zero-shot Prompting)}} \\
\hline
\textbf{Gemma-3} & \textbf{81.56} & \textbf{81.52} & \textbf{80.76} & \textbf{80.72} \\
Qwen2.5 & 80.65 & 79.15 & 79.62 & 77.83 \\
\hline
\multicolumn{5}{c}{\textit{Large Language Models (Few-shot Prompting)}} \\
\hline
Gemma-3 & 89.49 & 89.49 & 88.92 & 88.86 \\
\textbf{Qwen2.5} & \textbf{90.89} & \textbf{90.78} & \textbf{90.72} & \textbf{90.64} \\
\hline
\multicolumn{5}{c}{\textit{Large Language Models (Finetuning)}} \\
\hline
Gemma-2 & 84.11 & 82.46 & 83.74 & 81.80 \\
\hline
\end{tabular}
}
\end{table}

\section{Result Analysis and Discussion} \label{discussion}
To evaluate the effectiveness of NLI models on the proposed dataset, we conducted a comprehensive performance analysis across multiple data dimensions in Section~\ref{discussion}.

\subsection{Impact of Hypothesis Length}

\begin{figure}[ht]
    \centering
    \includegraphics[width=1\linewidth]{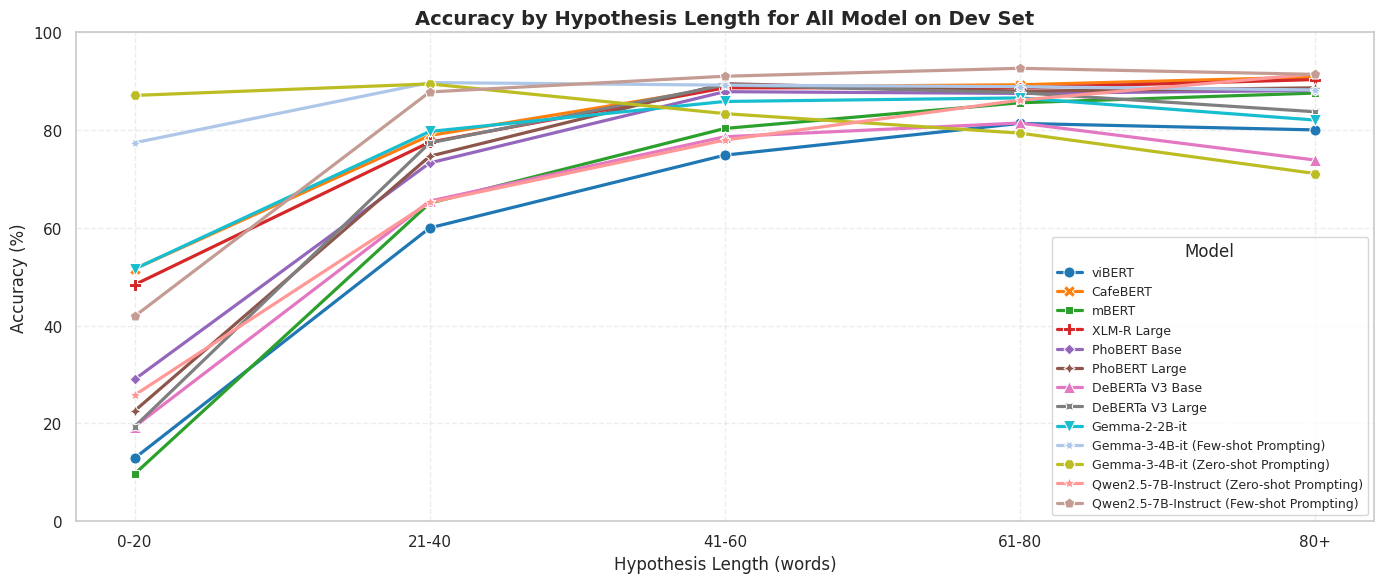}
\caption{Effect of hypothesis length on model accuracy.}
\label{fig:hyp_length_effect}
\end{figure}

To better understand how the amount of semantic information in hypotheses affects legal inference difficulty, we analyze model performance across different hypothesis lengths. Figure~\ref{fig:hyp_length_effect} illustrates the relationship between hypothesis length and model accuracy on the Dev set. Overall, performance improves substantially when hypothesis length increases from short (0–20 tokens) to medium (21–60 tokens), and then stabilizes or slightly declines for longer hypotheses.

Pretrained language models (PLM) such as viBERT \cite{VietviBERT2020}, mBERT \cite{Devlin2019}, and DeBERTa \cite{He2020DeBERTa} show relatively low performance on very short hypotheses, likely due to insufficient legal context. In contrast, instruction-tuned LLMs under few-shot prompting remain relatively robust, indicating stronger contextual reasoning ability. Medium-length hypotheses yield the best overall performance, suggesting that adequate semantic information—without excessive noise—is crucial for legal NLI reasoning.

\subsection{Impact of Lexical Overlap}

To examine whether NLI models rely on superficial lexical cues rather than genuine semantic reasoning, we analyze lexical overlap using multiple similarity measures.

\paragraph{Jaccard Similarity.}

\begin{figure}[ht]
    \centering
    \includegraphics[width=1\linewidth]{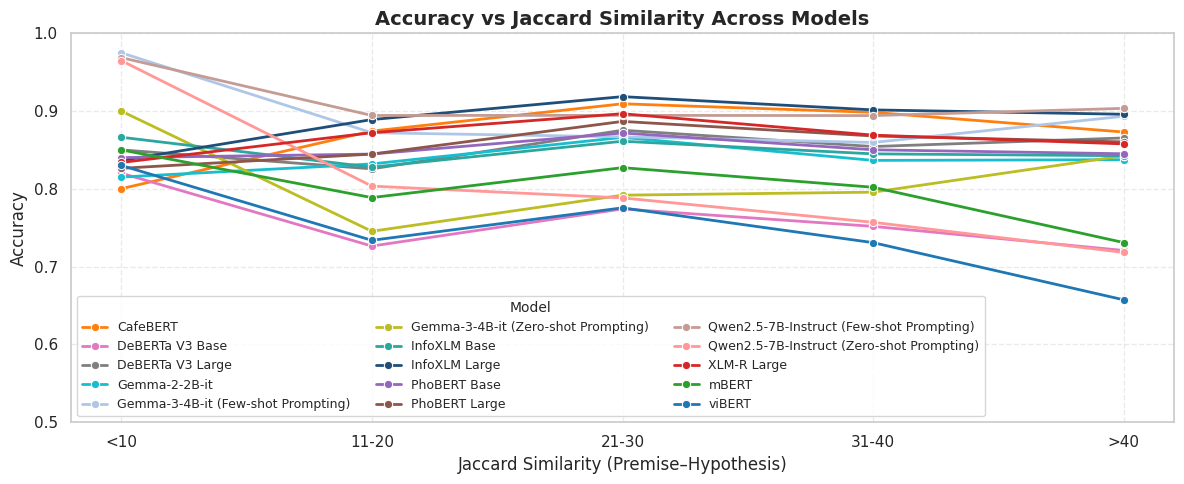}
\caption{Model performance by Jaccard similarity.}
\label{fig:jaccard_effect}
\end{figure}

Figure~\ref{fig:jaccard_effect} presents model accuracy across different levels of Jaccard similarity between premise and hypothesis. Accuracy does not increase monotonically with lexical overlap, indicating that models do not rely solely on surface word matching. Performance peaks at moderate overlap (11--30\%), where shared vocabulary provides useful semantic cues without causing spurious entailment. At high overlap levels (>40\%), smaller pretrained models exhibit performance degradation, likely due to over-reliance on superficial lexical similarity.

\paragraph{Longest Common Subsequence (LCS).}

\begin{figure}[ht]
    \centering
    \includegraphics[width=1\linewidth]{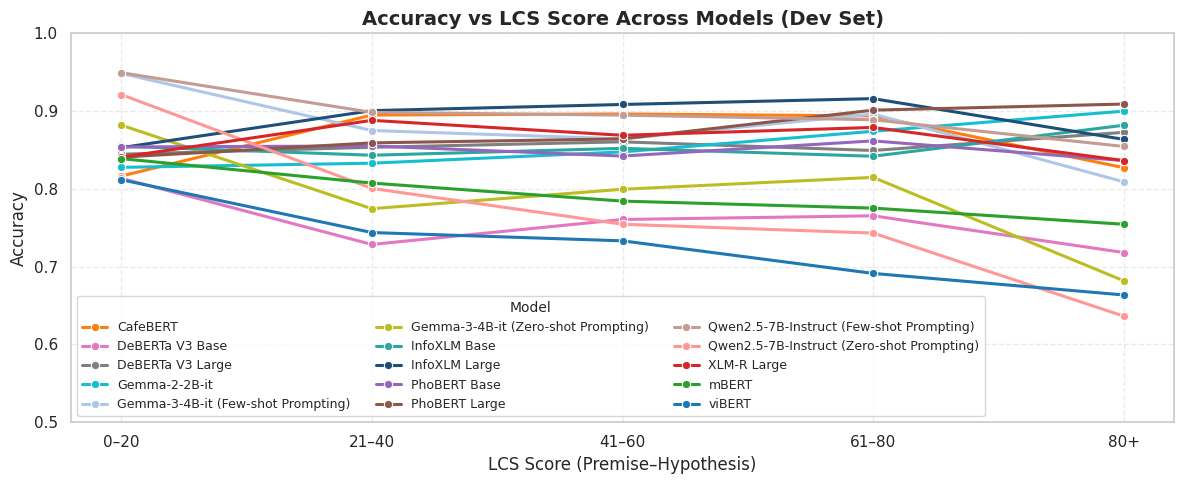}
\caption{Model performance across LCS ranges.}
\label{fig:lcs_effect}
\end{figure}

To further examine structural similarity, Figure~\ref{fig:lcs_effect} shows performance across different LCS  levels. Most models achieve optimal performance at moderate LCS levels (21--60 \%). Low LCS (0--20 \%) significantly reduces accuracy for traditional PLMs, reflecting limited deep semantic reasoning when structural similarity is minimal. In contrast, instruction-tuned LLMs maintain more stable performance across LCS ranges, demonstrating stronger generalization beyond surface structure.

\paragraph{New Word Rate.}

\begin{figure}[ht]
    \centering
    \includegraphics[width=1\linewidth]{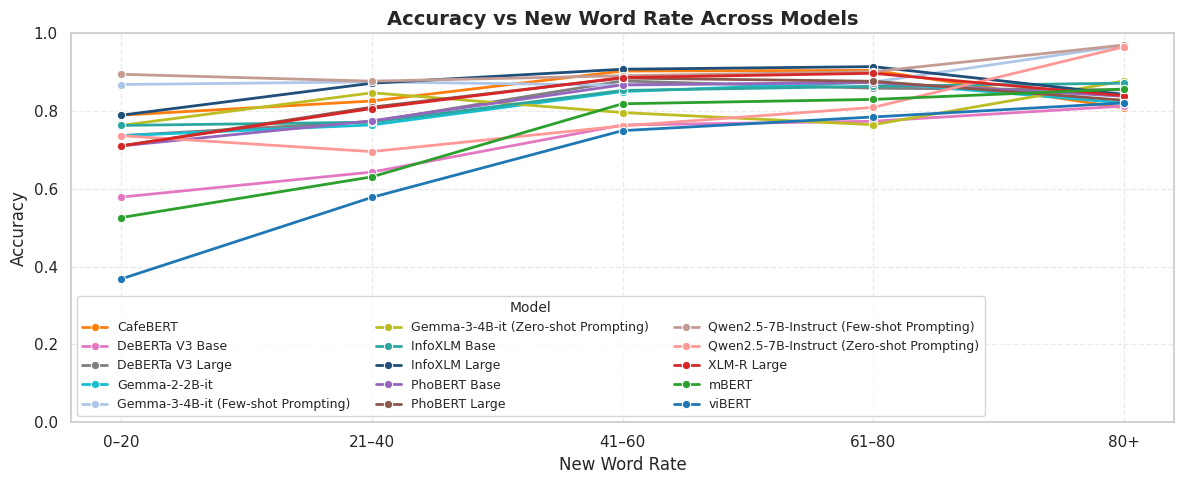}
\caption{Model performance by New Word Rate.}
\label{fig:newword_effect}
\end{figure}

Complementing the above analyses, Figure~\ref{fig:newword_effect} evaluates performance with respect to New Word Rate, defined as the proportion of hypothesis tokens absent from the premise. Accuracy generally improves from low to moderate levels (0--60 \%), suggesting that controlled semantic expansion enhances inference clarity. However, extremely high rates (80+ \%) may slightly degrade performance in fine-tuned PLMs, indicating difficulty in handling extensive semantic divergence. Instruction-tuned LLMs remain comparatively stable.

\subsection{Performance by Inference Label}

\begin{figure}[ht]
    \centering
    \includegraphics[width=0.9\linewidth]{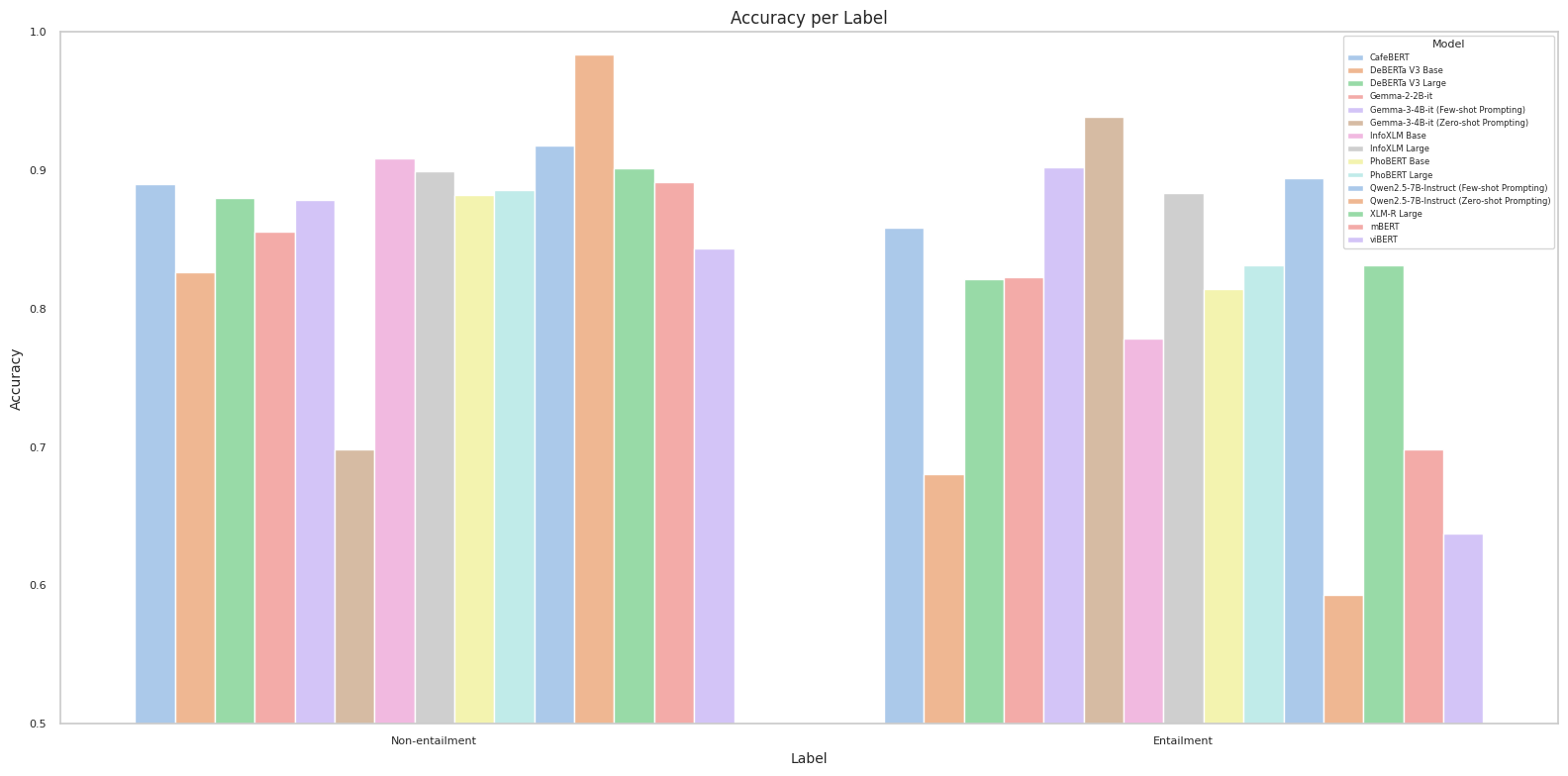}
    \caption{Accuracy by inference label.}
\label{fig:label_performance}
\end{figure}

To analyze differences in model performance across inference labels, we evaluate results separately for Entailment and Non-entailment instances. Figure~\ref{fig:label_performance} reports model accuracy for each label. 

Most models consistently achieve higher accuracy on Non-entailment than on Entailment. This suggests that Entailment is inherently more challenging, as it requires precise semantic inclusion and strict logical consistency, which are particularly demanding in legal texts.

This difficulty is more pronounced in traditional PLMs, which exhibit larger performance gaps between the two labels. In contrast, instruction-tuned LLMs (e.g., Qwen2.5-7B-Instruct and Gemma-3-4B-it) demonstrate more balanced and stable performance, especially under few-shot prompting, indicating stronger capability in handling fine-grained legal reasoning.

\subsection{Performance Across Legal Sub-domains}

The ViLegalNLI dataset is constructed from 168 legal documents spanning 27 domains, each characterized by distinct terminology and contextual structures. Analyzing performance across domains provides insights into the model’s ability to generalize to different legal areas.

\begin{figure}[ht]
    \centering
    \includegraphics[width=1\linewidth]{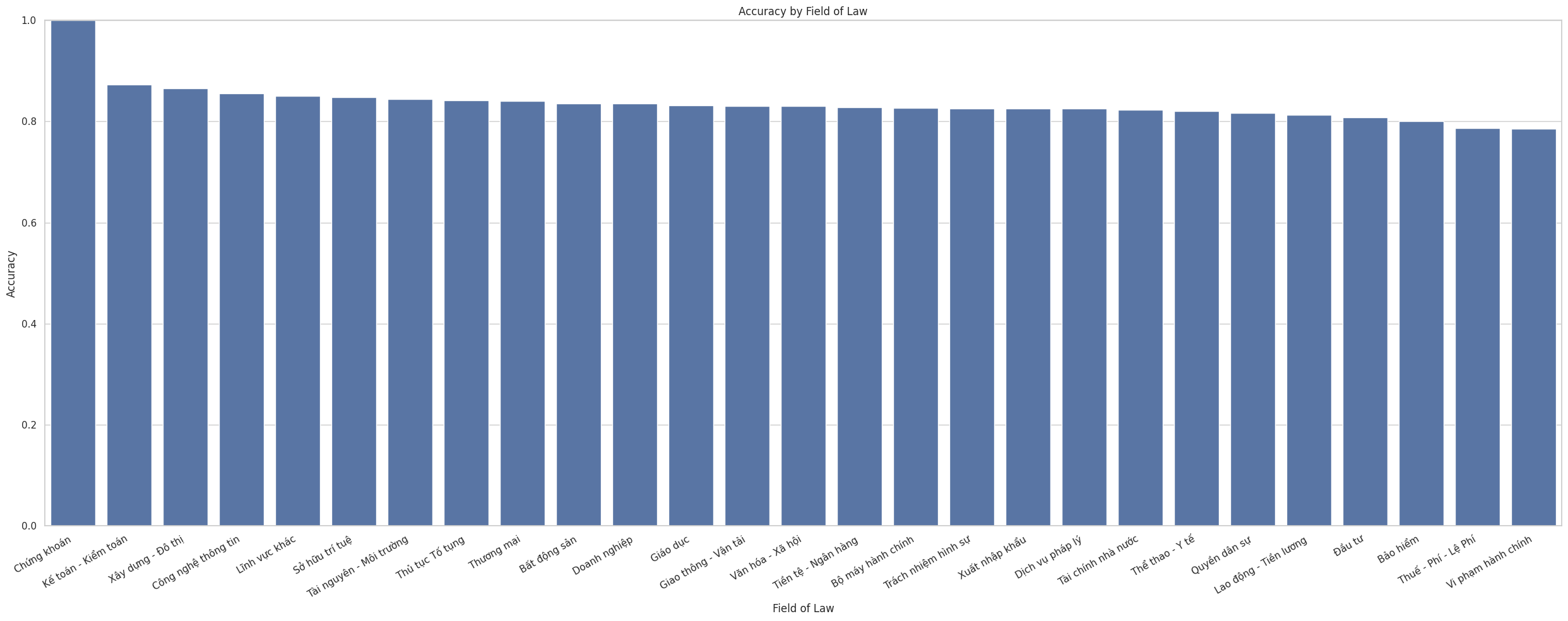}
\caption{Model accuracy across legal domains.}
\label{fig:domain_performance}
\end{figure}

Figure~\ref{fig:domain_performance} shows accuracy across 27 legal domains, revealing noticeable variation in performance. Higher accuracy is generally observed in sub-domains with more standardized language (e.g., civil, criminal, administrative law), whereas technically specialized domains (e.g., finance, taxation, securities, intellectual property) present greater challenges due to their complex terminology and domain-specific reasoning patterns.

\subsection{Impact of Generation Rules}

\begin{figure}[ht]
    \centering
    \includegraphics[width=1\linewidth]{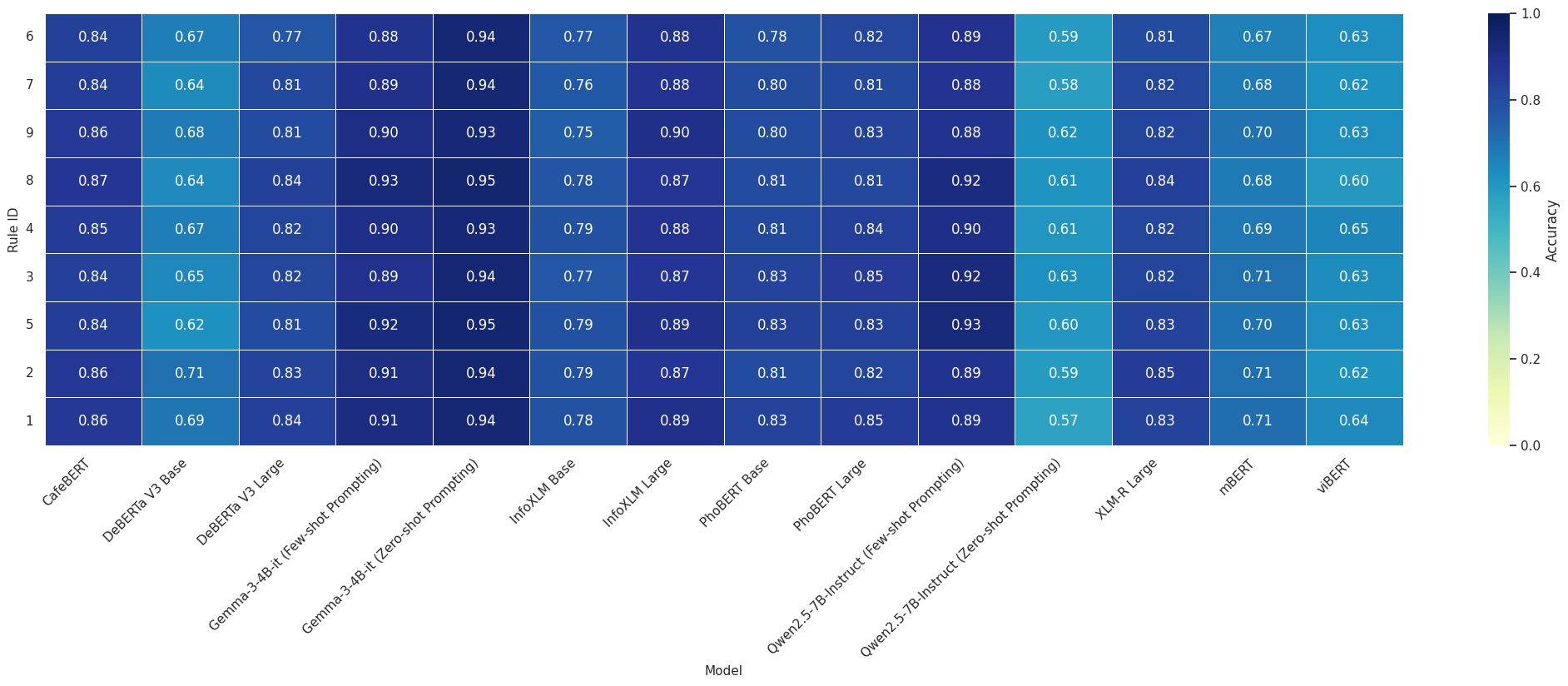}
    \caption{Performance across Entailment generation rules.}
\label{fig:ent_rule}
\end{figure}

\begin{figure}[ht]
    \centering
    \includegraphics[width=1\linewidth]{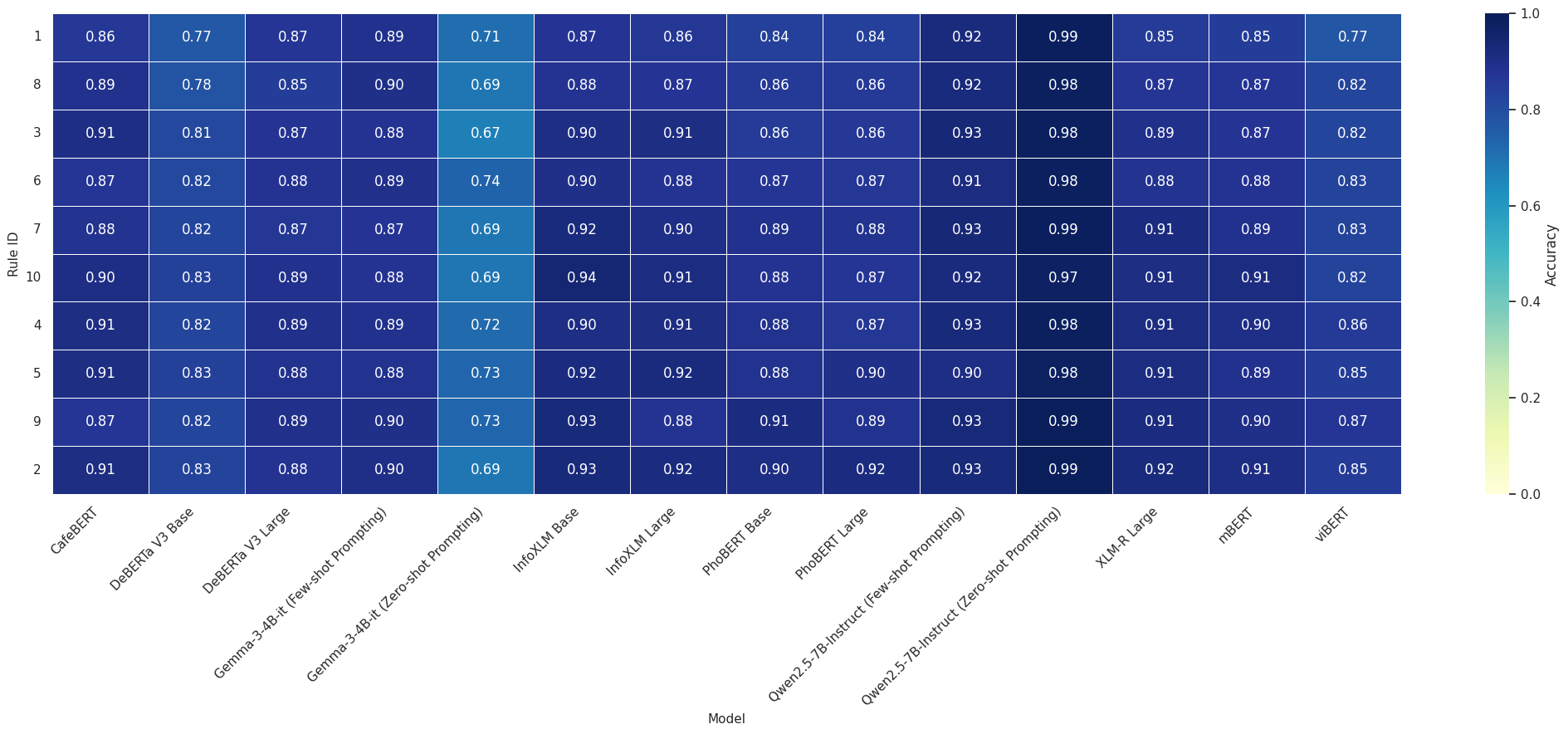}
\caption{Performance across Non-entailment generation rules.}
\label{fig:nonent_rule}
\end{figure}

To better understand model behavior, we analyze model performance across different data generation rules. This analysis allows us to examine how models handle diverse inference patterns (e.g., lexical substitution, negation, and semantic generalization) and to identify potential dataset artifacts or reasoning weaknesses. Figures~\ref{fig:ent_rule} and~\ref{fig:nonent_rule} illustrate model performance across hypothesis generation rules.

For Entailment, rules involving simple transformation that preserve core lefgal meaning tend to yield higher accuracy, whereas rules requiring deeper logical or causal reasoning lead to noticeable performance drops. In contrast, for Non-entailment, performance is generally higher and more stable, particularly when hypotheses exhibit clear semantic deviations from the premise. However, subtle logical inconsistencies remain challenging, indicating that models still struggle with fine-grained reasoning beyond explicit semantic differences.


\subsection{Cross-Domain Evaluation}

To assess domain generalization, we conduct cross-law evaluation, where the training, development, and test sets are drawn from different legal sub-domains. XLM-R (large) and CafeBERT are evaluated under both in-domain and cross-domain settings (Table~\ref{tab:cross_domain}).

The results indicate minimal performance degradation under cross-domain settings, suggesting that both multilingual and Vietnamese-specific models capture generalizable legal reasoning patterns. However, this robustness primarily reflects surface-level reasoning ability. While current models perform well under domain shifts, deeper legal inference remains challenging. Instruction-tuned LLMs exhibit stronger robustness and semantic generalization, highlighting their potential for Vietnamese legal NLI applications.

\begin{table}[ht]
\centering
\caption{Cross-domain evaluation results (\%).}
\label{tab:cross_domain}
\resizebox{\textwidth}{!}{
\begin{tabular}{lcccc}
\hline
\multirow{2}{*}{\textbf{Model}} & \multicolumn{2}{c}{\textbf{Dev set}} & \multicolumn{2}{c}{\textbf{Test set}} \\
\cline{2-5}
 & \textbf{Accuracy (\%)} & \textbf{F1-score (\%)} & \textbf{Accuracy (\%)} & \textbf{F1-score (\%)} \\
\hline
\multicolumn{5}{c}{\textit{In-domain evaluation results}} \\
\cline{2-5}
XLM-R (large) & 86.98 & 86.81 & 86.37 & 86.19 \\
CafeBERT & 87.56 & 87.44 & 87.49 & 87.36 \\
\hline
\multicolumn{5}{c}{\textit{Cross-domain evaluation results}} \\
\cline{2-5}
XLM-R (large) & 88.95 & 88.91 & 87.55 & 87.52 \\
CafeBERT & 88.39 & 88.39 & 87.98 & 87.96 \\
\hline
\end{tabular}
}
\end{table}

\subsection{Error Analysis}\label{sec:error_analysis}

\begin{figure}[ht]
    \centering
    
    \begin{subfigure}{0.48\linewidth}
        \centering
        \includegraphics[width=\linewidth]{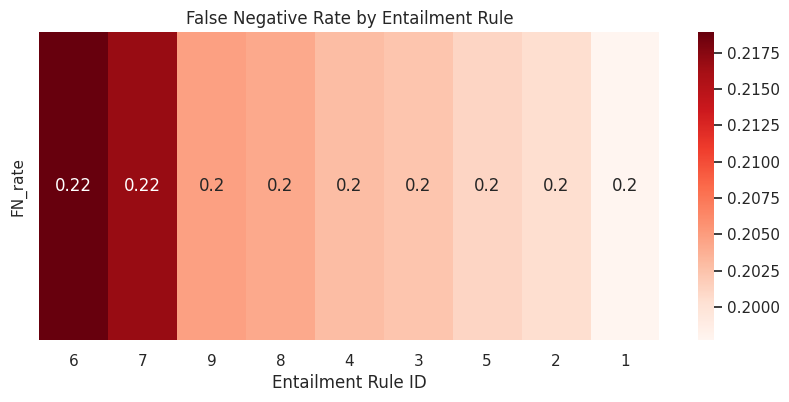}
        \caption{Entailment (False Negatives)}
        \label{fig:entailment_error}
    \end{subfigure}
    \hfill
    \begin{subfigure}{0.48\linewidth}
        \centering
        \includegraphics[width=\linewidth]{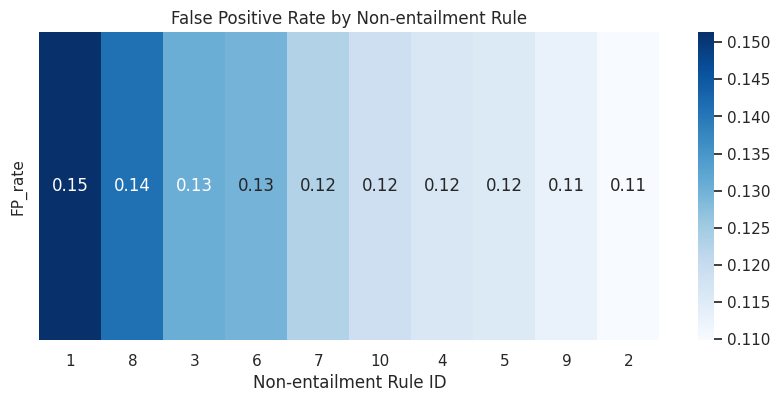}
        \caption{Non-entailment (False Positives)}
        \label{fig:non_entailment_error}
    \end{subfigure}
    
    \caption{Error analysis of prediction errors across entailment and non-entailment classes.}
    \label{fig:error_analysis}
\end{figure}

Although aggregate metrics such as Accuracy, Precision, Recall, and F1-score provide an overall assessment of model performance, they offer limited insight into the underlying causes of prediction failures. To better understand these limitations, we conduct a fine-grained error analysis by hypothesis type (Figure~\ref{fig:error_analysis}), aiming to uncover systematic error patterns and identify inference rules that pose particular challenges.

As shown in Table~\ref{tab:error_types}, the most prevalent error type is the misclassification of Entailment instances as Non-entailment (False Negatives), especially under Entailment Rules 6 and 7 (22\%). These rules require recognizing implicit clauses or legal consequences that are not explicitly articulated in the premise. The results suggest that models rely predominantly on surface-level semantic cues and struggle when entailment depends on implicit or consequence-based reasoning.

\begin{table}[ht]
\centering
\caption{Error Types and Their Proportions}
\label{tab:error_types}
\begin{tabularx}{\columnwidth}{p{2cm} p{2.5 cm} X c}
\toprule
\textbf{Type} & \textbf{Pattern} & \textbf{Description} & \textbf{Rate} \\
\midrule

Type 1: E $\rightarrow$ N
& Entailment Rules 6, 7
& Failure to recognize implied clauses and legal consequences derived from the premise.
& 22\% \\
\hline
Type 2: N $\rightarrow$ E
& Non-entailment Rule 1
& Being misled by lexical overlap.
& 15\% \\
\hline
Type 3: E $\rightarrow$ N
& Entailment Rules 8, 9
& Inability to reason from general principles to specific cases and to combine related legal provisions.
& 13\% \\
\hline
Type 4
& Model-specific Bias
& Tendency to predict Non-entailment when the model fails to understand the sentence meaning.
& 37\% \\

\bottomrule
\end{tabularx}
\end{table}

Entailment Rules 8 and 9 (13\%) further reinforce this limitation, as they involve multi-step inference and the integration of multiple related legal provisions. Such cases require hierarchical reasoning across clauses or articles, a capability that current models appear to handle inadequately. This finding underscores the inherent difficulty of modeling multi-layered logical dependencies in formal legal texts.

Conversely, False Positives primarily arise from Non-entailment Rule 1 (15\%), where substantial lexical overlap between premise and hypothesis misleads the model. In these instances, surface similarity is incorrectly interpreted as inferential validity. This behavior indicates insufficient separation between lexical resemblance and true semantic entailment.

\begin{figure}[ht]
    \centering
    \includegraphics[width=1\linewidth]{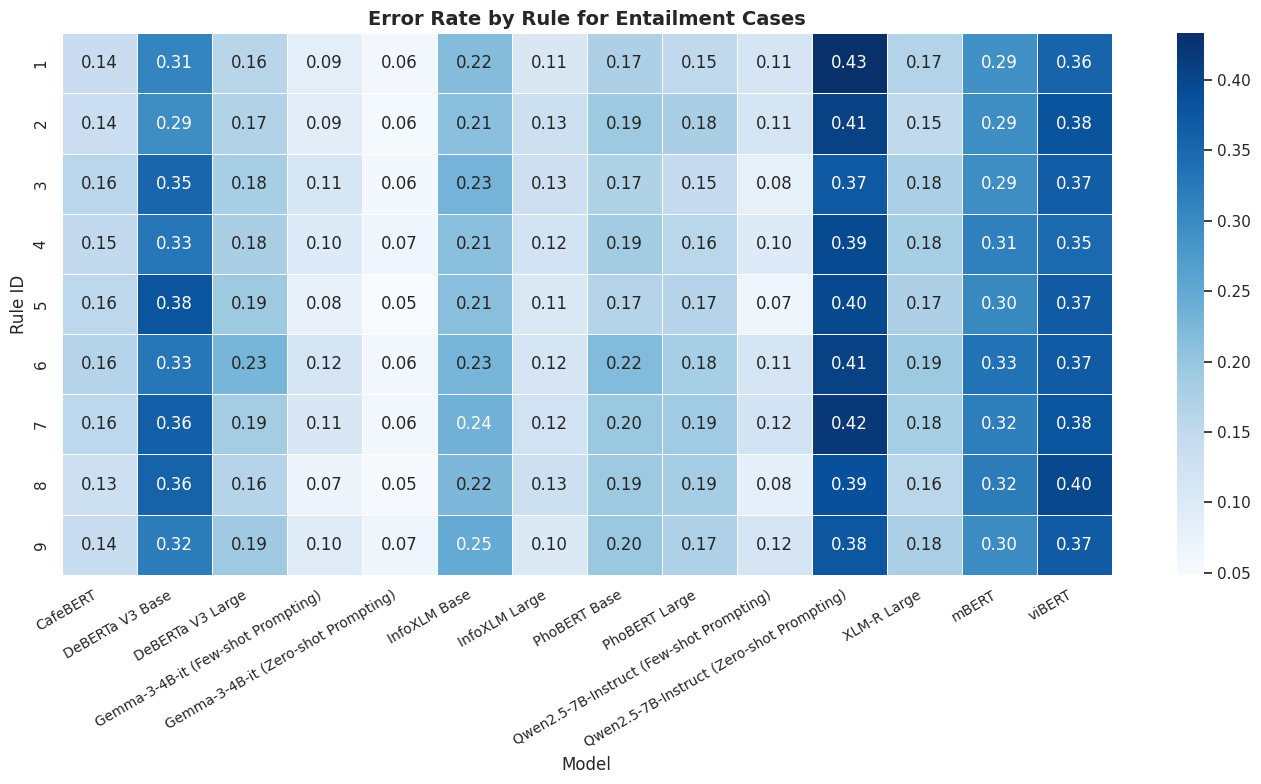}
    \caption{Error Rate by Generation Rule for the Entailment Class.}
    \label{fig:entailment_rule_error}
\end{figure}

Beyond rule-specific errors, the analysis also reveals a systematic prediction bias in large language models under zero-shot settings (Figure~\ref{fig:entailment_rule_error}). In particular, Qwen2.5 exhibits a strong tendency to predict Non-entailment when semantic interpretation is uncertain, leading to an error rate of up to 37\% in ambigous cases. This conservative prediction strategy suggests that, in the absence of task-specific guidance, the model defaults to a safer negative classification. 

This observation highlights the critical role of prompting design in improving reasoning reliability for legal NLI tasks and suggests that better calibration strategies may further reduce such biases.

\section{Conclusion and Future Work}\label{sec:conclusion_future_work}

This paper introduces ViLegalNLI, the first large-scale dataset for Vietnamese legal natural language inference, comprising 42,012 premise–hypothesis pairs derived from officially promulgated statutory documents. The dataset is constructed through a semi-automatic pipeline that leverages large language models for controlled hypothesis generation and initial labeling, followed by multi-stage validation to ensure consistency and reliability. ViLegalNLI serves as a standardized benchmark for evaluating Vietnamese legal NLI models.

We conduct comprehensive evaluations across multilingual pretrained models, Vietnamese-specific transformer models, and instruction-tuned large language models under both fine-tuning and in-context learning settings. The results demonstrate that domain-adapted models and few-shot LLM configurations achieve strong performance. However, our analysis reveals that current models still struggle with deeper legal reasoning, particularly in cases involving implicit inference, multi-step logic, and low lexical overlap.

In future work, we plan to extend the dataset to additional types of legal In future work, we plan to extend the dataset to additional types of legal documents and incorporate more complex inference phenomena, such as multi-step reasoning, exception handling, and cross-article dependencies. We also aim to expand the task to paragraph- and document-level inference, enabling more realistic modeling of legal reasoning in practical applications.


\section*{Declarations}

\textbf{Conflict of interest} The authors declare that they have no conflict of interest.

\section*{Data Availability}

Data will be made available on reasonable request.

\nocite{*}
\bibliography{sn-bibliography}

\end{document}